\documentclass[manuscript,screen,authorversion,nonacm]{acmart}

\AtBeginDocument{%
  \providecommand\BibTeX{{%
    \normalfont B\kern-0.5em{\scshape i\kern-0.25em b}\kern-0.8em\TeX}}}

\copyrightyear{2022} 
\acmYear{2022} 
\setcopyright{acmlicensed}
\usepackage{caption}
\usepackage{subcaption}
\usepackage{wrapfig}
\newcommand{\hbr}{HTE}

\begin{document}

% \title{Hierarchical Quality-Diversity for Online Damage Recovery}
%  \title{Hierarchical Quality-Diversity for Online Damage Recovery in the Physical World}
\title{Online Damage Recovery for Physical Robots with Hierarchical Quality-Diversity}


\author{Maxime Allard}
\affiliation{%
   \department{Adaptive and Intelligent Robotics Lab}
  \institution{Imperial College London \country{United Kingdom}}
%   \city{London}
%   \country{United Kingdom}
}
\email{maxime.allard@imperial.ac.uk}

\author{Simón C. Smith}
\affiliation{%
   \department{Adaptive and Intelligent Robotics Lab}
  \institution{Imperial College London \country{United Kingdom}}
%   \city{London}
%   \country{United Kingdom}
  }
% \email{s.smith-bize@imperial.ac.uk}

\author{Konstantinos Chatzilygeroudis}
\affiliation{%
\department{Computer Engineering and Informatics Department}
  \institution{University of Patras \country{Greece}}
%   \city{Patras}
%   \country{Greece}
  }
% \email{costashatz@upatras.gr}

\author{Bryan Lim}
\affiliation{%
  \department{Adaptive and Intelligent Robotics Lab}
  \institution{Imperial College London \country{United Kingdom}}
%   \city{London}
%   \country{United Kingdom}
  }
% \email{bryan.lim16@imperial.ac.uk}

\author{Antoine Cully}
\affiliation{%
  \department{Adaptive and Intelligent Robotics Lab}
  \institution{Imperial College London \country{United Kingdom}}
%   \city{London}
%   \country{United Kingdom}
  }
% \email{a.cully@imperial.ac.uk}

\renewcommand{\shortauthors}{Allard et al.}

\begin{abstract}
In real-world environments, robots need to be resilient to damages and robust to unforeseen scenarios. 
Quality-Diversity (QD) algorithms have been successfully used to make robots adapt to damages in seconds by leveraging a diverse set of learned skills.
A high diversity of skills increases the chances of a robot to succeed at overcoming new situations since there are more potential alternatives to solve a new task.
However, finding and storing a large behavioural diversity of multiple skills often leads to an increase in computational complexity. Furthermore, robot planning in a large skill space is an additional challenge that arises with an increased number of skills. Hierarchical structures can help reducing this search and storage complexity by breaking down skills into primitive skills. In this paper, we introduce the Hierarchical Trial and Error algorithm, which uses a hierarchical behavioural repertoire to learn diverse skills and leverages them to make the robot adapt quickly in the physical world.
We show that the hierarchical decomposition of skills enables the robot to learn more complex behaviours while keeping the learning of the repertoire tractable. Experiments with a hexapod robot show that our method solves a maze navigation tasks with 20\% less actions in simulation, and 43\% less actions in the physical world, for the most challenging scenarios than the best baselines while having 78\% less complete failures.

\end{abstract}


% \ccsdesc[500]{Computer systems organization~Robotics}



% \centering
% \includegraphics[width=1.0\textwidth]{Images/Paper_Arch_v4.png}
% \centering
% \caption{Schematic representation of a 3-layered Hierarchical Trial and Error repertoire for omni-directional hexapod locomotion. The top behavioural repertoire (BR) fetches three different solutions from the middle level BR which in turn fetches each leg controller for the robot individually to achieve the desired upper-level behaviour. The middle layer has a depth (represented by a cylinder) from which the secondary behaviour can be chosen.}
% \label{fig:hbr_architecture}

\maketitle

\begin{figure}
\centering
\includegraphics[width=1.0\textwidth]{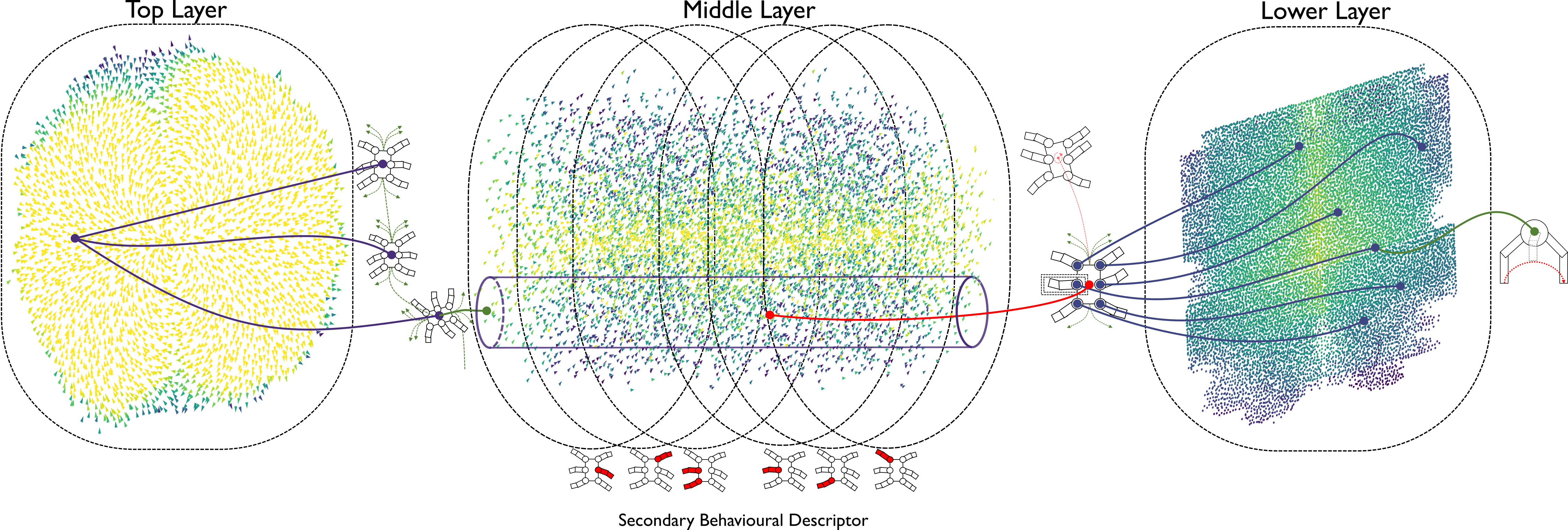}
\centering
\caption{Schematic representation of a 3-layered Hierarchical Trial and Error repertoire for omni-directional hexapod locomotion. The top behavioural repertoire (BR) fetches three different solutions from the middle level BR which in turn fetches each leg controller for the robot individually to achieve the desired upper-level behaviour. The middle layer has a depth (represented by a cylinder) from which the secondary behaviour can be chosen.}
\label{fig:hbr_architecture}
\end{figure}

\section{Introduction}

% Robot learning has significantly advanced in the last years~\cite{miki2022learning,akkaya2019solving,Cully} by using powerful learning methods, including Deep Learning, Reinforcement Learning and Quality-Diversity (QD)~\cite{shrestha2019review,sutton2018reinforcement,Long2016,Chatzilygeroudis2020Quality-DiversityOptimization}.
Robots can learn skills to solve various problems through methods coming from the field of robot learning~\cite{miki2022learning,akkaya2019solving,Cully}. Some of the learning methods include Deep Learning, Reinforcement Learning and Quality-Diversity (QD)~\cite{shrestha2019review,sutton2018reinforcement,Long2016,Chatzilygeroudis2020Quality-DiversityOptimization} to name a few.
However, the deployment of robots in real-world scenarios is still a hard problem~\cite{chatzilygeroudis2019survey}.
One of the main problems that robots need to tackle while being deployed in the physical world is the occurrence of unforeseen situations (such as damages) that can greatly alter the performance of the robot. Other problems might include robustness to external perturbations, generalisation of  experience in novel scenarios and interaction with other complex agents (both artificial and natural) among several others~\cite{thrun2002probabilistic}.

It is impossible for people developing robots to predict all the scenarios a robot will end up in and this makes it very difficult to train the robot for all possible situations. The literature proposes different techniques to counter this problem. For example, we can train an agent on relevant and automatically generated scenarios \cite{Fontaine2020,Gambi2019,Rocklage2017,Mullins2017} before the deployment or we can make the robot learn on how to adapt to unknown situations during deployment.
Instead of relying on a single fixed control policy, a robot that adapts has the potential to perform efficiently in previously unseen scenarios~\cite{aastrom2013adaptive}.
Adaptive control has been proposed as a way of tuning parameters during exploitation~\cite{aastrom2013adaptive,cameron1984self}. In autonomous robots, the parameters of the controller can be directly updated based on the error of a predictive model to maintain an exploratory behaviour~\cite{smith2020diamond, 9618829}. Another way to achieve adaptation is to leverage multiple diverse control policies. For example, methods that leverage Quality-Diversity (QD)~\cite{Long2016,7959075,Chatzilygeroudis2020Quality-DiversityOptimization} algorithms have shown that the diversity of pre-computed solutions is a key factor for fast adaptation~\cite{Cully,Chatzilygeroudis2018Reset-freeRecovery,Kaushik2020AdaptiveRobotics}.

Generating solutions beforehand requires a robot to have a lot of interactions with an environment. Having interactions with the physical world requires manual resets, hardware replacements, battery replacements (if no other power supply is provided) etc. Consequently, the learning process is usually done in simulation since it requires less supervision and time than running physical trials with a robot. All of these restrictions are not present in simulation which makes the simulation a less time-consuming and less laborious way of getting thousand hours of experience with a robot. Simulators, albeit being easier to use than the real world, are imperfect representations of the physical world which means that a simulation-to-real-world gap will always be present for the learned solutions. 
% To overcome the difference between simulated environments and the physical world in computer vision for example, some works propose to make the simulated environment as close as possible to the physical world by using Generative Adversarial Networks~\cite{Goodfellow2014} to make the simulation environment look like a real physical one~\cite{bousmalis2018}. Another approach is to apply domain randomisation~\cite{Sadeghi2016,Tobin2017} to the simulator by changing parameters (such as the friction coefficient, random forces, different initial positions etc.) of the simulation during the training process. The hope is that for the deployment in the physical world, the robot will have encountered at least one scenario that is close enough to reality in the simulator. 

% In addition to learning a diverse set of robotic skills, Quality-Diversity (QD)~\cite{Long2016,7959075,Chatzilygeroudis2020Quality-DiversityOptimization} algorithms have been used effectively to bridge the simulation-to-real-world gap by adapting learnt solutions with different methods such as Gaussian processes~\cite{Rasmussen2005GaussianLearning}. 
% They are used to find a repertoire with a large set of diverse and high-performing robotic skills. 

In the work of Cully et al.~\cite{Cully}, the authors present the Intelligent Trial and Error algorithm (ITE) which uses QD to find around 13,000 solutions for a hexapod robot to walk in different directions in simulation and then adapt to damages in seconds in the real-world. The training is done in simulation since each solution requires 5 seconds of walking. The training of the solutions was done with 40 million evaluations which corresponds to $40 \times 10^6 \times 5 = 2 \times 10^8$ seconds, or $6.3$ years of controller execution that would be necessary to train such a repertoire in the real-world. In a real-world scenario where the robot has a damaged leg, ITE uses its large repertoire with a diverse set of controllers to find the solutions that are unaffected by the leg damage. 
The algorithm updates a set of Gaussian processes~\cite{Rasmussen2005GaussianLearning} with experiences gathered by trial and error in the environment to find the best solution in the pre-trained repertoire. 
In ITE, the repertoire of solutions is used to solve a single skill. To accomplish a task such as navigation under damage, the algorithm requires more skills, e.g. turning left, turning right, moving backward, etc. However, it is intractable to generate all the required solutions for each of the skills.
Other approaches to repertoire-based adaptation with QD include the Reset-Free Trial and Error algorithm (RTE) ~\cite{Chatzilygeroudis2018Reset-freeRecovery} and Adaptive Prior Selection for Repertoire-Based Online Learning algorithm (APROL) \cite{Kaushik2020AdaptiveRobotics}.
RTE and APROL generate diverse controllers that are used for planning in order to solve navigation tasks with a damaged robot. APROL takes advantage of multiple trained repertoires with different prior knowledge on possible future conditions, e.g. damage or change in the friction between the robot and the ground, whereas RTE only uses a single repertoire.

All of the approaches above assume that a subset of the pre-trained solutions in a repertoire can be used effectively to achieve adaptation in the real world.
The stochasticity of the environment and the complexity of the tasks may break this assumption if the diversity of solutions is not large enough.
One approach to tackle this problem is to increase the number of solutions in the repertoire, but at the cost of sampling efficiency, memory storage and a larger solution space. Increasing the number of stored solutions might increase the diversity but it will make it harder to select the correct skill. An increased solution space can impact the training procedure since it will make it more difficult to fully explore and optimise the space. Furthermore, any algorithm that needs to select the learned skills for downstream tasks (e.g. path planning algorithm) will have to chose among thousands of solutions which makes the task even more complex.
% Another approach is to use  hierarchies for the organisation of different levels of behavioural abstraction.

To tackle these issues, it is possible to increase the diversity of solutions for robot control by decomposing the search-space as a hierarchy and efficiently organising controllers~\cite{Merel2019}. The different layers allow for compositionality and re-usability of solutions (i.e. controllers). Furthermore, the use of hierarchies can help with transferability to other domains and few-shot learning~\cite{eppe2020hierarchical,Cully2018,Etcheverry}.
% In this work, we propose to leverage a hierarchical structure of repertoires with controllers to increase the diversity of solutions and improve the overall adaptability of a robot.
In this work, we propose to combine the quality of finding diverse solutions coming from QD and the efficiency of hierarchical structures by learning hierarchical controllers to make robots adapt quicker to unforeseen scenarios.

To this end, we introduce the Hierarchical Trial and Error (HTE) algorithm ,which uses hierarchical behavioural repertoires (HBRs) \cite{Cully2018}. HBRs are composed of layers with different repertoires of solutions. Each repertoire stores solutions that range from low-level motor commands to high-level task goals descriptions. In HBRs, the layers are connected by one or more edges, implemented as the genotype of each solution and each layer executes one or more behaviours from the layers below it.
The different layers of \hbr{} are trained to exhibit different behaviours on the robot (e.g. moving a single leg, walk for 1 second, etc). 
% By including new behavioural descriptor dimensions, i.e. a secondary behavioural descriptors in the layers of \hbr{}, it is possible to increase the diversity of the solutions during the deployment.
Following RTE, our method uses the experience generated by the robot during the deployment phase to update a set of Gaussian processes. The Gaussian processes are used in a planning step with an MCTS algorithm to find the best action in the hierarchical repertoire based on the current state of the robot and the goal of the task. 
% The hierarchical organisation of controllers and introduction of secondary behavioural descriptors allows our method to find diverse solutions in various unseen scenarios efficiently. 

To test our hierarchical algorithm, we compare it first to RTE and APROL in a set of experiments with a hexapod robot in a simulated maze environment before evaluating our approach in the real world. For the real world experiments, we build a similar maze to the one in simulation and remove a leg from the hexapod robot to test how well we can adapt in real conditions. Additionally, the robot is never trained with damage nor does it train in the physical world.

Finally, we want to answer the following questions around the usefulness of \hbr{}:
\begin{itemize}
\item Is it possible to find diverse and useful controllers for omni-directional hexapod locomotion with \hbr{}?
% \item How does the diversity compare to vanilla QD algorithms trained on the same behavioural space?
\item Can we increase the diversity of learnt skills by using an additional secondary behavioural descriptors in the hierarchy?
% \item Is it helpful to chose the secondary behavioural descriptors carefully?
\item Can \hbr{} solve complex downstream robot tasks both in simulation and in the physical world?
\item Is \hbr{} able to adapt to damages in real-world experiments?
\item Can we leverage the diversity of skills while keeping the execution time of \hbr{} tractable?
\end{itemize}

Our results show that \hbr{} can find diverse solutions by decomposing the search space hierarchically. This makes the optimisation process simpler in comparison to finding solutions in the full extended solution space.
In comparison to the baselines, \hbr{} requires less steps to solve the task and less failures in a maze navigation task (i.e. downstream task) since it can effectively leverage the large collection of hierarchical skills.

\begin{figure}%
\centering
     \begin{subfigure}[b]{0.45\textwidth}
         \centering
         \includegraphics[width=1.0\textwidth]{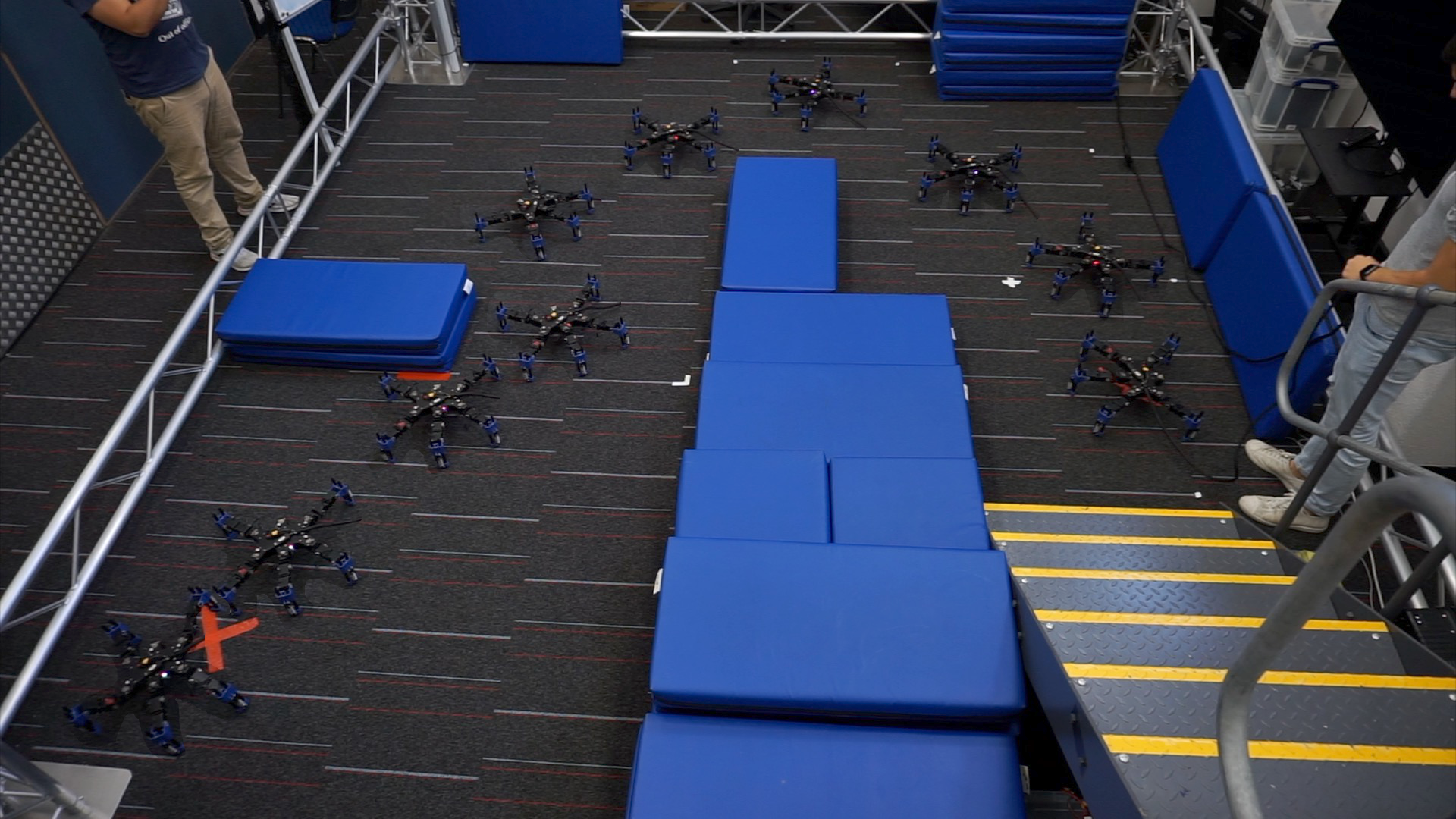}
         \caption{Adaptation in a Maze Environment}
         \label{fig:real_maze}
     \end{subfigure}
     \hfill
     \begin{subfigure}[b]{0.45\textwidth}
         \centering
         \includegraphics[width=0.85\textwidth]{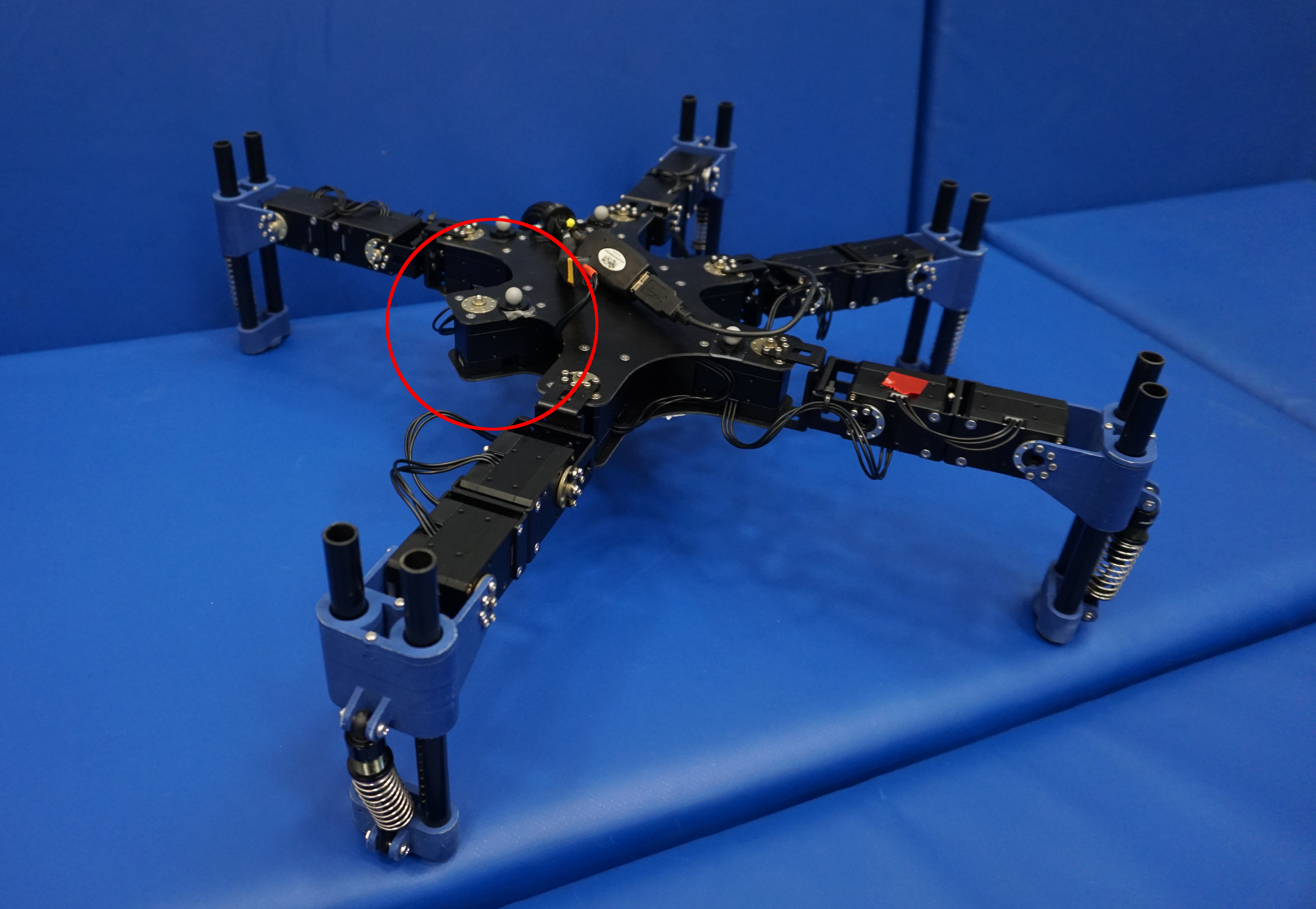}
         \caption{Hexapod robot with a damaged leg (red circle)}
         \label{fig:real_hexa}
     \end{subfigure}
     \caption{Experimental Setup For the Physical Downstream Experiments}
\end{figure}

% BREAKS THE ANONYMITY, INCLUDE UPON PUBLICATION
This work extends the original "Hierarchical Quality-Diversity for Online Damage Recovery" \cite{Allard2022} paper in three aspects. 
First, we show that the Trial and Error process for the selection of secondary behaviours, compared to a random selection, helps the hexapod robot to adapt faster in the case of damages.
Second, we explore the capacity of the algorithm to perform well in the physical world by running experiments with a real robot. The Hierarchical Trial and Error algorithm is able to adapt quicker to damages in the real-world in comparison to the baselines. This result is specially relevant as adaptation is harder in the physical world than in simulation. Furthermore, solutions learned in simulation might not translate directly to the physical world. This gap is reduced by the adaptation and online optimisation mechanisms of the algorithm.
Lastly, the hierarchical architecture reduces the adaptation time. To showcase the reduction in adaptation time we compare different algorithms in the physical world and record their time to complete the maze task.

\section{Background and Related work}

% \textit{\textbf{Quality-Diversity.}}
\paragraph{\textbf{Quality-Diversity.}}
Quality-Diversity has shown to be a powerful optimisation tool that can solve complex problems such as robot damage recovery fast and reliably~\cite{Cully}, achieving state-of-the-art results on unsolved Reinforcement Learning tasks in sparse-reward environments \cite{Ecoffet2021FirstExplore}, human-robot interaction~\cite{fontaine2020quality}, robotics~\cite{nordmoen2021map,Grillotti2021UnsupervisedOptimisation,eysenbach2018diversity,Mouret2020QualityOptimization,salehi2021br}, games~\cite{perez2021generating,sarkar2021generating,steckel2021illuminating} or constrained optimisation ~\cite{fioravanzomap} among several others.

Quality-Diversity (QD) algorithms ~\cite{Long2016,Chatzilygeroudis2020Quality-DiversityOptimization,7959075} are a family of evolutionary algorithms that aim to generate a collection of diverse and locally high-performing solutions. The diversity of solutions is obtained by defining a \emph{behavioural descriptor} $\mathbf{b}$ (also known as feature vector)  used to characterise a solution. This diversity of solutions contrasts with classical optimisation algorithms that only look for the highest performing solution. For example, a mobile agent can reach a final destination by following different (diverse) paths. Each one of these paths is a valid solution to reach the target but they are all define different trajectories to reach the destination.
In QD algorithms, the expert defines the $n-$dimensional $\emph{behavioural\ space}$ as $B \in \mathbb{R}^n$. After evaluation, each solution $\theta$ is associated with a behavioural descriptor $\mathbf{b_\theta}$ that defines their behaviour in the space $B$.
Usually, the engineer defines the behavioural descriptor $\mathbf{b_\theta}$ manually, which requires expert knowledge and can result in a solution bias during the evolutionary process. To define the behavioural descriptor automatically, recent methods encode the behaviours into a latent space with dimensionality-reduction algorithms ~\cite{Cully2018,Grillotti2021UnsupervisedOptimisation,Cully2019,PaoloUnsupervisedSpace,Laversanne-FinotCuriositySpaces} to discover behaviours in the latent space.

The solutions $\theta$ are defined by a genotype $\mathbf{g}$, which belongs to the  $k$-dimensional $\emph{genotype\ space}\ G \in \mathbb{R}^k$. 
The most popular QD algorithms include MAP-Elites~\cite{Mouret2015} and Novelty Search~\cite{Lehman2011}. We use MAP-Elites in our method to generate the repertoires.
In MAP-Elites, once a solution has been evaluated, it is stored in a grid-like archive, called repertoire. The repertoire discretises the $\emph{behavioural\ space}\ B$ to store individuals in distinct cells.
In the case that two solutions have the same behavioural descriptor, the algorithm keeps the one with the highest fitness in the repertoire and discards the other.

% \textit{\textbf{Hierarchical Organisation of Autonomous Control.}}
\paragraph{\textbf{Hierarchical Organisation of Autonomous Control.}}
In nature, the nervous system of complex organisms, e.g. mammals, use hierarchical controllers for robust and versatile behaviours \cite{Merel2019}. Even more, authors in cognitive psychology have deemed hierarchies as critical mechanisms for developing intelligent agents~\cite{eppe2020hierarchical}.

Hierarchical structures allow algorithms to compose complex solutions out of primitive ones, which helps with the optimisation process.
For example, hierarchies are used to stack different levels of abstracted goals to solve high-dimensional problems in Reinforcement Learning ~\cite{Gehring2021HierarchicalExploration,Nachum2018} or to enable hierarchical latent spaces to discover a diversity of behavioural representations in evolutionary algorithms~\cite{Etcheverry}.

Furthermore, Hierarchical Genetic Algorithms \cite{Ryan2020Pyramid:Algorithms} use hierarchies with fitness functions of different granularities per layer, allowing the evolution of both coarse and fine behaviours at different levels.
Also, hierarchical abstractions are effective at controlling robots in more complex environments.
Several works show that the hierarchical abstraction of behaviours allows robots to solve complex tasks ~\cite{Duarte2016,Jain2020FromLocomotion,Li2021PlanningLocomotion}. In these works, a neural network uses planning to choose the actions from a repertoire of  primitive behaviours. 

Quality-Diversity has been successfully combined with hierarchical structures for a rapid divergent search of solutions with trees \cite{Smith2016} or to create hierarchical behavioural repertoires (HBR) that can be transferred across different types of robots to solve complex tasks such as drawing digits with a robotic arm \cite{Cully2018}. Our method is inspired by this latter work on HBRs and reuses the concept to create a diverse set of solutions. 
% Similarly to mammals, we aim to use information factorisation \cite{Merel2019} in our hierarchy to create efficient subsystems that control a robot.

% It is important to note that the majority of the improvements gained by the use of hierarchies comes at the cost of expert knowledge and design decision for the hierarchies to be a useful abstraction of the task, agent and environment.

% This removes all influence from hand-engineered behaviours and lets the algorithm discover them by itself.\\

% \textit{\textbf{QD-Based Adaptation.}}
\paragraph{\textbf{QD-Based Adaptation.}}
% Adaptive robot control is defined by both a controller that can modify its parameters and a method to modify them during execution~\cite{aastrom2013adaptive}.
Adaptability is a key feature for controlling robots in realistic scenarios. Usually, real-world scenarios define intractable state and solution spaces. It is impossible for a robot to have a pre-defined solution for each possible situation. Thus, the robot has to adapt to any occurring situation in real-time. 
% Traditional adaptive techniques include self-tuning of parameters, e.g. online PID controller adaptation~\cite{cameron1984self} and adaptation of internal models~\cite{datta1996adaptive}. Modern approaches include self-organisation of behaviour as an online adaptation to the environment of the robot~\cite{smith2020diamond}. 
% Moreover, robot adaptation has successfully been achieved by leveraging the diversity of high performing solutions generated by QD algorithms and Bayesian optimisation.

% A meta-evolution fitness function is presented Bossens et al.  In this work, the overall capacity of adaptation of the solutions is part of the evolutionary process~\cite{bossenshal02555231}.
%In games, adaptive QD techniques have been used to update the difficulty of the task presented to the player~\cite{gonzalez2020finding} and for opponents to adapt to their partner's strategy~\cite{canaan2020generating}.
Cully et al. introduced the Intelligent Trial and Error (ITE) algorithm as an adaptive control method based on MAP-Elites and Gaussian processes for planning and adaptation~\cite{Cully}. The most significant results of ITE include the rapid adaptation of a hexapod robot with damaged legs. Some other works build upon ITE to introduce a local adaptation mechanism to improve the simulation-to-reality transfer ~\cite{kim2019exploration}, to optimise a swarm of robots ~\cite{bossens2020rapidly} or to find new game levels with different difficulties \cite{Duque2020}. 
Along the same lines, Chatzilygeroudis et al.\cite{Chatzilygeroudis2018Reset-freeRecovery} introduced the  Reset-free Trial-and-Error (RTE) algorithm. In this algorithm, a robot is able to adapt to unseen scenarios, e.g. damage in a leg, while executing a navigation task. The algorithm uses an repertoire of solutions and Gaussian processes to quickly find well performing solutions for this task. We base our work on this method and introduce further details in the next section.
Similarly, the Adaptive Prior Selection for Repertoire-Based Online Learning algorithm (APROL)~\cite{Kaushik2020AdaptiveRobotics}, adapts the behaviour of a hexapod to different damages while executing a navigation task. To do so, the algorithm chooses the best skill, using Gaussian processes, among a set of dozens of repertoires of solutions that have been trained on potential scenarios (e.g. damage to the legs or different friction coefficients) that the robot could face during deployment. In an iterative process, APROL selects the best solutions from the repertoire that is the most likely to represent the actual conditions (i.e. friction coefficient or leg damage). 
These methods show that the solutions created by QD are useful to adapt to different situations while executing a task. Similarly, the Hierarchical Trial and Error algorithm is built upon the idea to create a diverse set of solutions with an HBR first and subsequently find the best skill for a situation with Bayesian Optimisation.
% Swarm map-based Bayesian optimisation  allows for a swarm of robots to adapt to ever-changing resources in the environment with both centralised and decentralised optimisation methods~\cite{bossens2020rapidly}. 

\section{Preliminaries and Motivation}

\subsection{Reset-free Trial and Error}
The Reset-free Trial and Error algorithm (RTE) ~\cite{Chatzilygeroudis2018Reset-freeRecovery} is a powerful method to enable a robot to adapt to changing environments or scenarios in real time. First, a repertoire of solutions is created by MAP-Elites in simulation where each solution $\theta$ is stored with respect to the observed behavioural descriptor $\mathbf{b}_\theta$. 
% This results in a function mapping the behavioural descriptor to a relative outcome  $f: \mathbf{b}_\theta \rightarrow O$.
After creating the repertoire in simulation, it is used as a prior for the mean of a set of Gaussian processes. The main loop of RTE consists in running a Monte Carlo Tree Search (MCTS)~\cite{remi2006mcts} algorithm to plan the next best action $\mathbf{b}_{t+1}$ together with the Gaussian processes given the actual state of the robot $s_t$. Since the simulation environment is always imperfect, Gaussian processes map the behavioural descriptors $\mathbf{b}_\theta$ to observed behaviours $\mathbf{b}_{observed}$ in the new environment. Once the action from the repertoire is executed, the observed behavioural descriptors are used to update the Gaussian process before RTE does the next round of planning.
% These Gaussian processes map the $\mathbf{b}_\theta$ descriptor to the effective observed outcome $\mathbf{b}_{observed}$ of a controller when that controller is executed in the current situation. 
Finally, this main loop of RTE runs until a stop criterion is met, e.g. reaching the goal state. 
% For each RTE episode, a MCTS planning algorithm is used together with the Gaussian processes for selecting the behaviour descriptor $b_{t+1}$ of the next action to be performed based on the actual state of the robot $s_t$. 
% The action is executed by the robot and the outcome is observed. The Gaussian processes are then updated with this new experience $f(b_{t+1})$.

RTE has been successfully used to enable a hexapod robot and a velocity-controlled differential drive robot to reach its final destination even with one or more damages. The algorithm takes advantage of the diversity of skills created by the MAP-Elites algorithms and shows to be very effective at finding solutions in the repertoire of diverse solutions that are suitable for different situations. In contrast to ITE, RTE has more diverse skills but only one way to execute them. This behaviour implies that in an unexpected situation the algorithm needs to have a viable solution in the repertoire for each skill which is not always the case. For this reason, an effective algorithm relies on a good diversity of skills with a redundancy of ways to execute these skills. On one hand, ITE promotes the diversity of executing a single skill and on the other hand RTE pushes for a diversity of skills without a diverse way of executing them. Our method aims to optimise for both properties to make the adaption of robots more effective.

% During the MAP-Elites optimisation process, the experience of the agent in the environment is mapped as a function of actions to relative outcomes $f: A \rightarrow O$.

% The most common method to initialise $P_k$ in RTE is with MAP-Elites. MAP-Elites is able to generate the required individuals $i$.
%  This experience is stored in a set $D^d_{1:t} = {f_d(a_1),\ldots,f(a_t)}$ and used as prior to fit the initial Gaussian processes $GP_d$ for each dimension $d$ of $O$.
% After this initialisation, the main loop of RTE runs until a stop criterion is met, e.g. reaching the goal state. For each RTE episode at time $t$, an MCTS (Monte Carlo tree search) planning algorithm is used for selecting the next action $a_{+1}$ based on the actual state of the robot $s_t$. The action is executed by the robot and the outcome is observed. The new experience $f(a_{t+1})$ is added to $D$ and all the Gaussian processes are updated.

% RTE has been successfully tested in navigation and manipulation tasks. For example, a hexapod robot is able to reach its final destination even with a damaged leg. Also, a  velocity-controlled differential drive robot is able to navigate a maze when damage is applied to its configuration. Even with these positive results, generalisation to more complex scenarios is not trivial. Results of RTE can still be improved in terms of convergence (scenarios where the goal of the task is not found) and speed (reduce the number of actions taken to find a solution) (\red{I think we need quantitative arguments here}).

\subsection{Hierarchical Behavioural Repertoires}\label{hbr_back}

To increase the number of ways to achieve different skills, \hbr{} extends the Hierarchical Behavioural Repertoire (HBR) algorithm \cite{Cully2018} to learn locomotion skills for robots. HBR algorithms show to be effective at finding complex skills that can be used across different robots (e.g. drawing digits with a robot-arm and a humanoid robot \cite{Cully2018}) without retraining all the layers. 
Structurally, HBRs consist of layers of MAP-Elites repertoires which are chained together to create diverse skills. A bottom layer could   consist of a controller that makes a robotic arm move to any reachable point in a defined space. Following this architecture, we can train a second layer that draws a line from one point to another by reusing solutions from the first layer (i.e. reach two points with the arm). These new skills can be used by a third layer to create arcs with five different lines. Finally, the robotic arm can learn how to draw digits by drawing different arcs with the skills from the third layer.
All the layers are trained sequentially (i.e. one after the other) from the bottom up since each repertoire of behaviours (i.e. layer) is using previous behavioural repertoires to create more complex skills. In the special case of only using one layer, the HBR algorithm reduces to a classical QD algorithm with a single repertoire. HBR algorithms are able to create complex solutions (e.g. draw digits with a robotic arm) by building on top of previously built repertoires. \hbr{} uses the HBR algorithm to optimise its skills in different sub spaces (i.e. different behavioural repertoires) of the full solution-space for the complex skills it needs to adapt during the deployment.

% \textit{\textbf{Behavioural Descriptor and Genotype.}}
\paragraph{\textbf{Behavioural Descriptor and Genotype.}}
Each layer $k$ in \hbr{}, contains solutions $\theta_k$ with a behavioural descriptor $\mathbf{b_k}$. The solutions are defined by their genotype $\mathbf{g}_k$ where $\mathbf{g} \in G$. In the following section, $k=1$ is the lowest layer which directly interfaces with the robot whereas $k={2,\dots K}$ are subsequent layers for a $K$ layered hierarchical behavioural repertoire.
Since multiple layers are stacked on top of each other, each repertoire needs to be connected to the others. In HBRs, this is done by using a genotype $\mathbf{g}_k$ definition that maps to the behavioural space of the other layers.

% \textit{\textbf{Stacking.}}
\paragraph{\textbf{Stacking.}}
The solutions $\theta_1$ in the lowest layer correspond to the parameters we use to directly control a robot. For the other layers $k$, the solutions $\theta_k$ in layer $k$ use the behavioural descriptor space of the lower layers as the action space. In practice, this is implemented as a succession of behavioural descriptor coordinates $\mathbf{b}_{k-1}$ which is defined in the genotype $\mathbf{g}_k$ of a solution. This definition allows to produce a successive execution of the corresponding controllers in the lower layers $k-1$. 
In other terms, the middle layer is a mapping $\phi():B_2 \rightarrow B_1$ from the behavioural descriptor space of the middle layer to an output space, which corresponds to the behavioural descriptor space of the lower layer.

Hierarchical repertoires are not limited by the number of layers and can create increasingly complex solutions by simply stacking more layers of repertoires. 

\section{Hierarchical Trial And Error}
\label{sec:methods}

\subsection{Hierarchical Architecture}
Our Hierarchical Trial and Error (\hbr{}) algorithm leverages the hierarchical repertoires to improve the diversity of skills, decompose complex solution spaces and make the adaption to different damage scenarios more effective. 
In this implementation, the \hbr{} algorithm uses a three-layered HBR (see Fig.~\ref{fig:hbr_architecture}) where (i) the bottom layer defines the movement for a single leg for 1 second, (ii) the middle layer makes the robot walk for 1 second by selecting 6 controllers for the legs and (iii) the top layer makes the robot walk for 3 seconds by chaining 3 controllers from the middle layer. Even though our implementation has three layers, \hbr{} can be used with an arbitrary number of layers in the hierarchical behavioural repertoires and it can be naturally applied to different tasks/robots.

% \textit{\textbf{The \hbr{} Architecture.}}\label{sec:arch}
\paragraph{\textbf{The \hbr{} Architecture.}}\label{sec:arch}
With the \hbr{} algorithm, our goal is to find behaviours that control the robot in various directions while doing it in many different ways. 
To this end, the \textbf{bottom} layer is directly controlling the legs of the hexapod. A solution in that repertoire is an open-loop controller for one single leg. 
% The hexapod robot has 18 servo-motors and each motor is controlled by a periodic function $\varphi(t,a,p,d)$ at each time-step $t$ with parameters $a$ for the amplitude, $p$ for the phase shift and $d$ for duty-cycle.
Each leg of the hexapod has three motors, where the motor attached to the body uses a periodic function $\varphi_1(t,a,p,d)$ at each time-step $t$ with parameters $a$ for the amplitude, $p$ for the phase shift and $d$ for the duty-cycle (see Cully et al.~\cite{Cully} for more information on the duty-cycle and the periodic function). The two other motors use the same periodic function $\varphi_2(t,a,p,d)$ such that the legs are always perpendicular to the bottom.

We use $a_1,p_1,d_1$ to control the first motor and $a_2,p_2,d_2$ to control the second and third motors. This configuration means that the \textit{genotype} consists of 6 parameters namely, $g_1=\{a_1,p_1,d_1,a_2,p_2,d_2\}$. 
The \textit{behavioural descriptor} $\textbf{b}_{l}$ for the bottom layer is measured at the end of a simulation of 1 second for each leg. The descriptor is defined as the height $h$ of the move, the swing distance $d_{swing}$ and the duty cycle $d$ of the second and third motor. The duty-cycle $d$ in this case is both a parameter and a \textit{behavioural descriptor}.
To encourage minimal energy consumption, the \textit{fitness function} is set as the sum of the commands that are sent to the motors throughout the simulation.

For the \textbf{middle} layer, the \textit{genotype} consists of selecting 6 leg behaviours from the bottom layer $g_2=\{\textbf{b}_{1},\textbf{b}_{2},\dots,\textbf{b}_{6}\}$ where $\textbf{b}_{l}$ is a behavioural descriptor vector for solutions of the bottom layer from a specific leg where $l \in \{1,2,3,4,5,6\}$.
The \textit{behavioural descriptor} has 3 dimensions and consists of the $x$, $y$ and $yaw$ displacement of the robot after one second.
The \textit{fitness function} measures the angular distance between the robot's final orientation and an ideal circular trajectory (similar to existing works~\cite{Cully2013}). The resulting repertoire consists of controllers that enable the robot to walk in any direction for one second.

Lastly, in the \textbf{top} layer, the \textit{genotype} consists of 3 behavioural descriptors from the middle layer, $g_3=\{\textbf{b}_1,\dots,\textbf{b}_3\}$. 
The \textit{behavioural descriptor} of this top layer is the final $x$, $y$ position (i.e. $x$, $y$ displacement, similar to Cully and Mouret~\cite{Cully2013}) of the robot which is measured after 3 seconds of simulation. 
%For this repertoire, we use a grid that divides each dimension of the behavioural space in 100 discrete cells.
Our \textit{fitness function} is similar to the middle layer namely the angular distance between a perfect trajectory and the robot's orientation.

\begin{table}
    \centering
    \begin{tabular}{p{2cm}|c || c || c}
    \toprule
                                 & Top Layer & Middle Layer & Lower Layer  \\
    \midrule
        Discretisation      & 100x100 Grid & 0.05 l-value & 0.01 l-value \\
        Mutation Rate       & 0.14   & 0.11 & 0.17  \\
        Generations         & 20000 & 30000 & 5001 \\
        Genotype Size       & 9 & 18 & 6\\
    \bottomrule
    \end{tabular}
    \caption{QD Parameters used for \hbr{} with a population of 200 individuals. The mutation is polynomial with $\eta_m$ and $\eta_c$ both at 10.0. The discretisation of the behavioural space is defined by a grid (Top Layer) or an $l$-value. }
    \label{tab:qd_params}
\end{table}

% \textit{\textbf{Primary and Secondary Behavioural Descriptors.}}\label{cond_behaviours}
\paragraph{\textbf{Primary and Secondary Behavioural Descriptors.}}\label{cond_behaviours}
To find a larger diversity of solutions in QD algorithms, we could make the behavioural space $B$ larger by adding additional dimensions to it. However, this can quickly lead to an exponential growth of the number of stored solutions in the repertoire which will make it i) challenging to cover the behavioural space due to the decrease of evolutionary selection pressure that happens with large collections \cite{7959075} and ii) difficult to store it in memory even on modern computers. In hierarchical repertoires, if the dimensionality of the behavioural space $B$ on lower levels increases, we can achieve more combinations for the high level skills (and thus increase the diversity). The difference between both approaches, is that lower layers in hierarchical repertoires can store fewer and simpler solutions to cover an equivalent behavioural space than a traditional single-layered, or "flat", repertoire. To be more descriptive, we can take the example of robot locomotion. Let's assume a robot can walk to many different positions in 3 seconds, then we would have to store all the solutions that could get us to different positions (i.e. $x$,$y$ displacement is our behavioural descriptor) if we are using a single-layered repertoire. However, in a hierarchical setting we can store solutions that get us to different positions in 1 second (i.e. a much smaller behavioural space) and then combine these solutions to get 3 second locomotion skills. The combination of skills doesn't require any additional storing of solutions and is thus more efficient. This means that the growth in size is significantly lower than it is for complex solutions that are trained in a traditional repertoire.
% This hierarchical decomposition of the behavioural space helps to avoid the expansion of the behavioural descriptor space $B$ found in Vanilla QD algorithms and improves exploration of the sub-spaces of the space $B$ across the hierarchy. This can be connected to the idea of \textbf{information factorisation} \cite{Merel2019}, where each sub-system requires only partial information to work.
To distribute different information flows to different subsystems in the hierarchy (i.e. information factorisation \cite{Merel2019}) in HBRs we can increase the size of our behavioural space $B$ in the middle layer of the HBR by increasing $B$ with additional \textbf{secondary} behavioural descriptor $\mathbf{b}_{\textrm{sec}}$ dimensions. The dimensions of the behavioural descriptor which were originally described in the previous section are called \textbf{primary} dimensions and they remain unchanged. The additional dimensions of our behavioural space are \textbf{secondary} which means that higher level layers do not have to define these as part of their genotype when selecting solutions from that lower repertoire. The secondary dimensions can be used to execute solutions with the same \textbf{primary} behavioural descriptors but in different ways. It is important to note that \textbf{secondary} behavioural dimensions are not used during the training of the higher-level layers but only used to keep a diversity of skills that can be used during the adaption phase. If we would simply add these dimensions to the middle layer during the training process, we would increase the search space for the higher levels which is something we want to avoid. Adding the dimensions to the top level layer would result in a very large behavioural space which in turn decreases evolutionary pressue as mentioned in the previous paragraph.

The training of the upper layers remains unchanged by directly selecting the solution in the lower layer with the nearest primary behavioural descriptor from the genotype value, without considering the secondary descriptor. Doing so, reduces the search space for the top level and helps the optimisation process to find diverse skills.

In our implementation of \hbr{} for the hexapod, we extend our middle layer with 6 additional dimensions for the secondary descriptors. In addition to the 3 existing primary dimensions $\textbf{b}_{\textrm{prim}}$, we now have the secondary behavioural descriptor values $\textbf{b}_{\textrm{sec}}$ which measure the ground contact for each leg (1 if it is in contact with the ground for more than 30\% of the time, otherwise 0). Since we have 6 legs, this gives us 6 additional secondary dimensions with 64 different possibilities (see Fig.~\ref{fig:hbr_architecture}). The top layer will keep the same genotype as it focuses only on \textbf{primary} behavioural descriptors during the training. The \textbf{secondary} descriptors will be used by \hbr{} during the adaptation phase to select the most appropriate way to execute each skill given the condition of the robot. 
In Fig.~\ref{fig:conditional_behaviours_grid}, we can see the additional diversity of the \textbf{secondary} behaviours in the middle layer by executing the top layer skills for different secondary descriptors.

\begin{figure}
\centering
\includegraphics[width=0.6\textwidth]{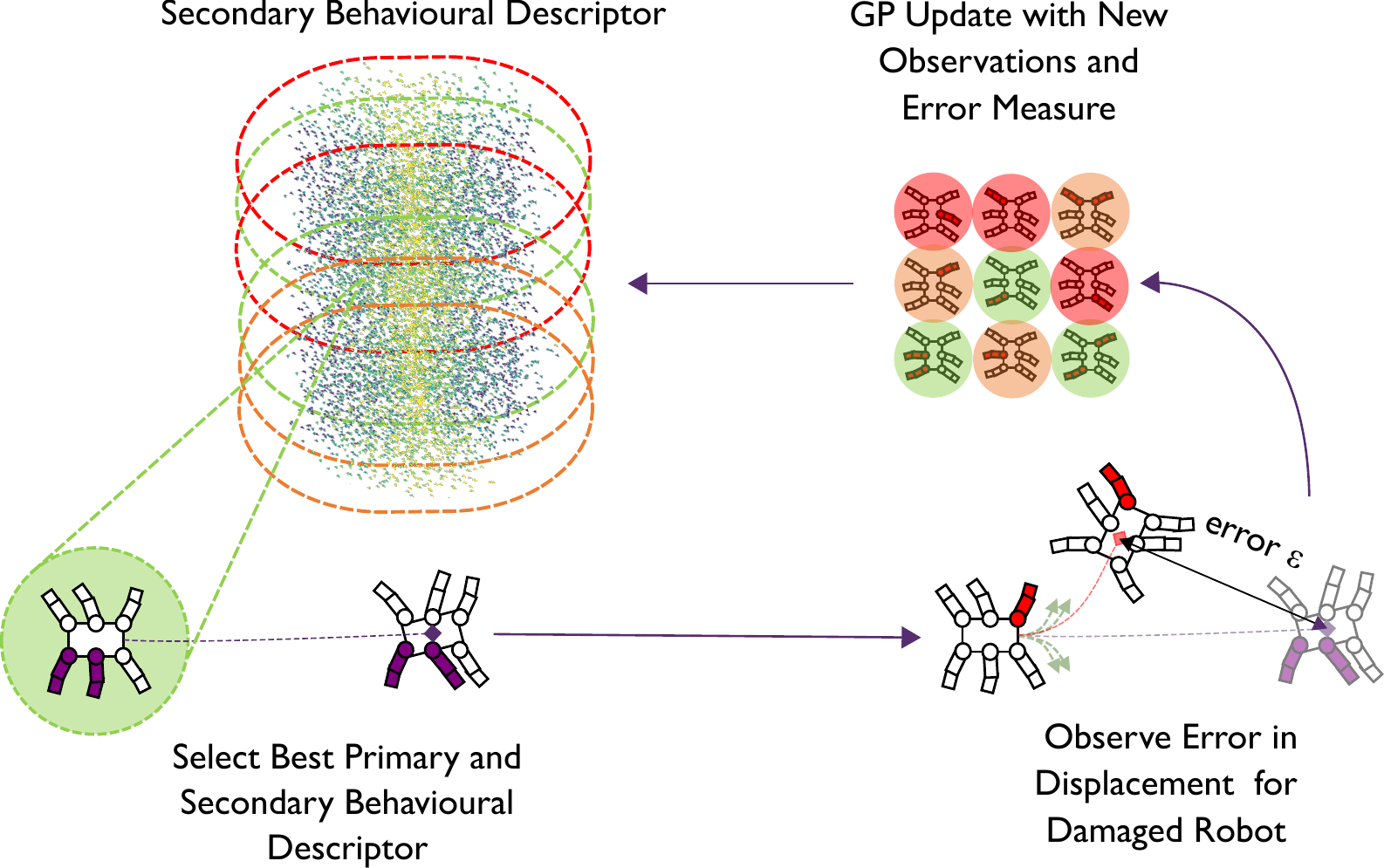}
\centering
\caption{Trial and Error Selection of Secondary Behavioural Descriptors. First, we select a secondary descriptor (purple leg) and a primary descriptor ($x$ and $y$ displacement) and execute it. Second, we observe the error of behaviour on the damaged robot (red leg) and lastly we update the Gaussian processes to find secondary descriptors that match the observed behaviour.}
\label{fig:gp-ucb}
\end{figure}

\subsection{Online Hierarchical Skill Adaptation}
We are interested in leveraging the diversity that we have incorporated into the hierarchical architecture to adapt the behaviour of the robot to different situations in the context of a maze navigation task. The adaption is done by selecting skills that are able to recover their behaviour for a specific situation (e.g. damages). 

% To find these unaffected skills, we have to select them amongst all the behaviours in our behaviour space which is now composed out top level behaviours and the additional secondary descriptors.
% \textit{\textbf{Primary Behaviour Selection via RTE.}}
\paragraph{\textbf{Primary Behaviour Selection via RTE.}}
HTE combines the adaptation and planning capabilities of RTE with the skill expressivity of the HBR. To this end, RTE uses the HBR in \hbr{} like a traditional single-layered repertoire by exclusively interacting with the top layer. RTE chooses which skill should be executed from the repertoire to come closer to the next goal by using MCTS and GPs. Since \hbr{} can execute each high-level skill with different secondary behavioural descriptors, we do not only need to choose which high-level skill to execute with RTE but we also need to select a secondary behavioural descriptor to adapt to the situation. 
%  Note that our behavioural descriptor is always between 0 and 1 which is why we use $l$ to centre the score around 0. In our experiments we usually centre this value with $l=0.5$ since a behavioural descriptor of $0.5$ means that there is no movement. Since we know the desired behaviours of the robot (i.e. $\mathbf{p}_{desired}$) from our trained repertoire we can simply calculate the error with respect to the real observed behaviour. 
 
%  \textit{\textbf{Secondary Behaviour Selection via Trial and Error.}}\label{sec:trial}
\paragraph{\textbf{Secondary Behaviour Selection via Trial and Error.}}\label{sec:trial}
To find the optimal secondary descriptor and adapt to an unforeseen situation, \hbr{} minimises minimises the error $\varepsilon$ between the observed primary behavioural descriptor and the desired one. We define $\varepsilon$ as::
 
% \begin{equation}
% \mathbf{\epsilon} = \exp{(-k\frac{\delta}{2|\mathbf{b}_\theta|-c})}
% \label{eq:error}
% \end{equation}

\begin{align}
\delta&=|\mathbf{b}^\prime-\mathbf{b}_\theta|\\
%\begin{split}
\mathbf{\varepsilon}&=\exp{(-k\frac{\delta}{2|\mathbf{b}_\theta|-c})} \label{eq:error}
%\end{split}
\end{align}
 
where $\delta$ is the error between the observed primary behavioural descriptor $\mathbf{b}^\prime$ and desired primary behavioural descriptor $\mathbf{b}_\theta$. Both $\mathbf{b}^\prime$ and  $\mathbf{b}_\theta$, are the behavioural descriptor vectors with the $x$, $y$ coordinates of the robot's centre of mass and the desired yaw rotation  ($\mathbf{b} = \{x, y, yaw\}$) (see Sec.~\ref{sec:arch}), and $k,c$ are  positive hyper-parameters to scale and shift the data respectively. In our setting we used $c=0.5$ and $k=4$. This error measure is inspired from the error measure introduced by APROL~\cite{Kaushik2020AdaptiveRobotics}.

To minimise the error $\varepsilon$ (Eq.~\ref{eq:error}), we use Bayesian Optimisation with Gaussian processes (GPs)~\cite{Rasmussen2005GaussianLearning}. GPs are a family of stochastic processes that are particularly attractive for this kind of regression problems because of their data-efficiency and uncertainty quantification. The GPs will learn to predict the error $\delta$ that a desired primary behaviour will have at every step for all combinations of a high-level skill and a secondary behaviour. Finally, to select the optimal secondary descriptor, we will use the upper-confidence bound (UCB) acquisition function~\cite{Srinivas2009GaussianDesign}. The UCB method will pick the secondary descriptors that minimise the error $\varepsilon$ given the current situation of the robot. This process is done during the deployment of the robot in the new situation without any resets (see Fig. \ref{fig:gp-ucb}). At each iteration of the trial and error iteration in \hbr{}, the GPs are updated with the new observed behaviours to refine their predictions of the least affected secondary descriptors.

In \hbr{}, we have a finite set of possible secondary behaviours and thus we can iterate over them to find the best high-level skill with RTE and the best secondary behavioural descriptor for a situation with UCB.
Note that adaptation is not only useful for any unforeseen situation (e.g. damage) but it can improve the simulation-to-real-world transfer of skills which is important in robotics.

% This means that the robot will be able to explore which secondary behaviours result in the same behaviours that were encountered during the training process while trying to solve a task. (e.g. solving a navigation task with RTE). 
% Figure \ref{fig:gp-ucb} depicts the overall process of selecting the secondary behavioural descriptor while interacting with an environment.

% \begin{figure}[h]
% \centering
% \includegraphics[width=0.5\textwidth]{Images/Maze.pdf}
% \centering
% \caption{Maze Navigation Task. The damaged (left middle leg) hexapod robot needs to reach the purple circle in a least amount of steps.}
% \label{fig:environment}
% \end{figure}

\section{Experimental Evaluations in Simulation}
 \subsection{Experimental Setup}
 
\begin{wrapfigure}{i}{0.5\textwidth}
  \begin{center}
    \includegraphics[width=0.35\textwidth]{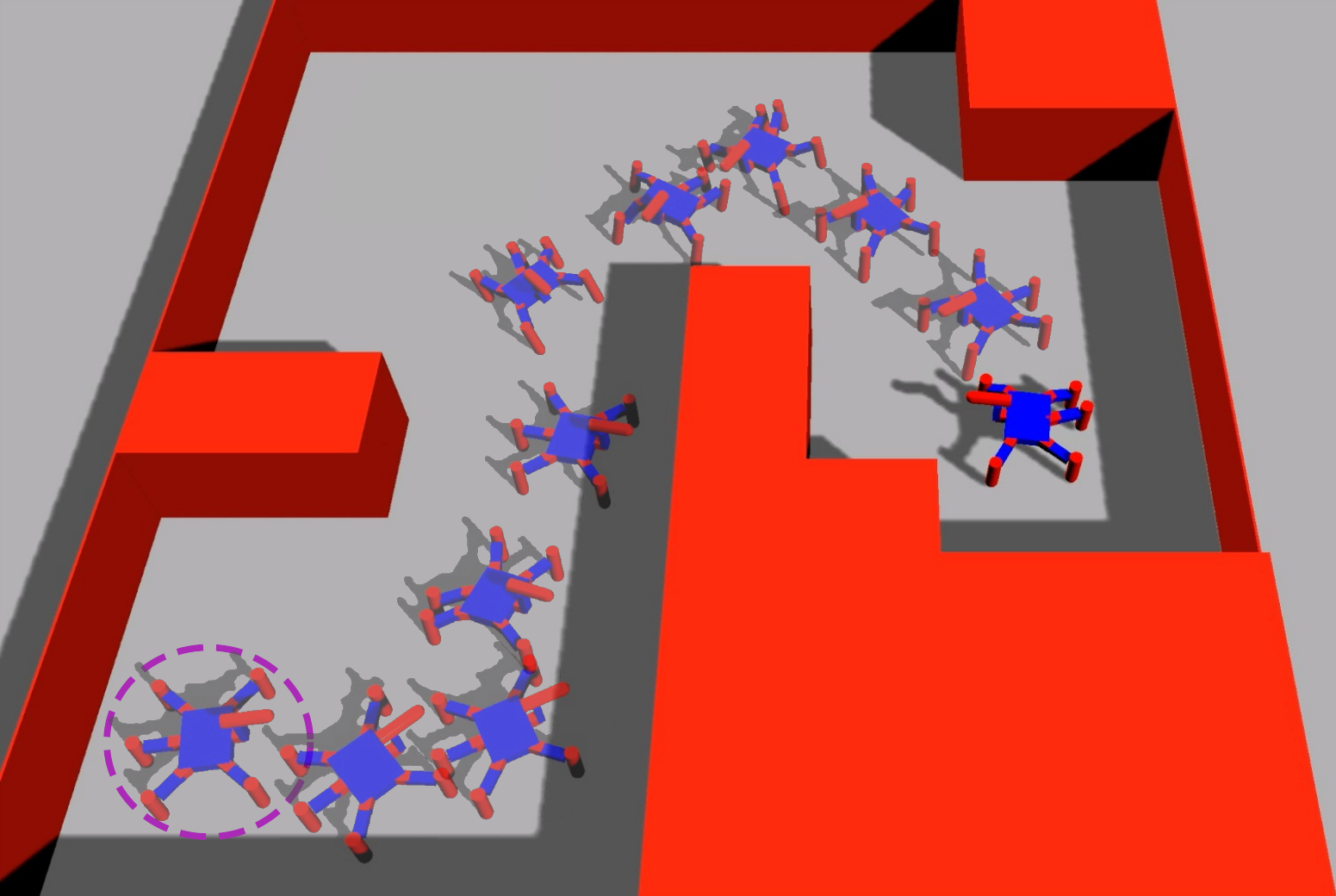}
  \end{center}
  \caption{Maze Navigation Task. The damaged (left middle leg) hexapod robot needs to reach the purple circle in a least amount of steps.}
\label{fig:environment}
\end{wrapfigure}

\begin{wrapfigure}{o}{0.2\textwidth}
  \begin{center}
    \includegraphics[width=0.15\textwidth]{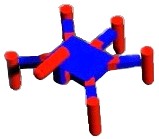}
  \end{center}
  \caption{Hexapod robot with the middle leg that as been blocked in the air to simulate a physical damage.}
\label{fig:hexa_damaged}
\end{wrapfigure}

 We evaluate the benefits of \hbr{} (with parameters in Table \ref{tab:qd_params}) in a maze solving task (see Fig. \ref{fig:environment}) with the hexapod being damaged in several ways. 
 We create 7 different scenarios where a different damage is present within each: 6 scenarios where one of the legs is blocked in the air (different leg for each scenario) as shown in Fig.~\ref{fig:hexa_damaged} and one scenario where the two middle legs are damaged at the same time.
 
\begin{figure}%[h!]%
\centering
\includegraphics[width=0.8\textwidth]{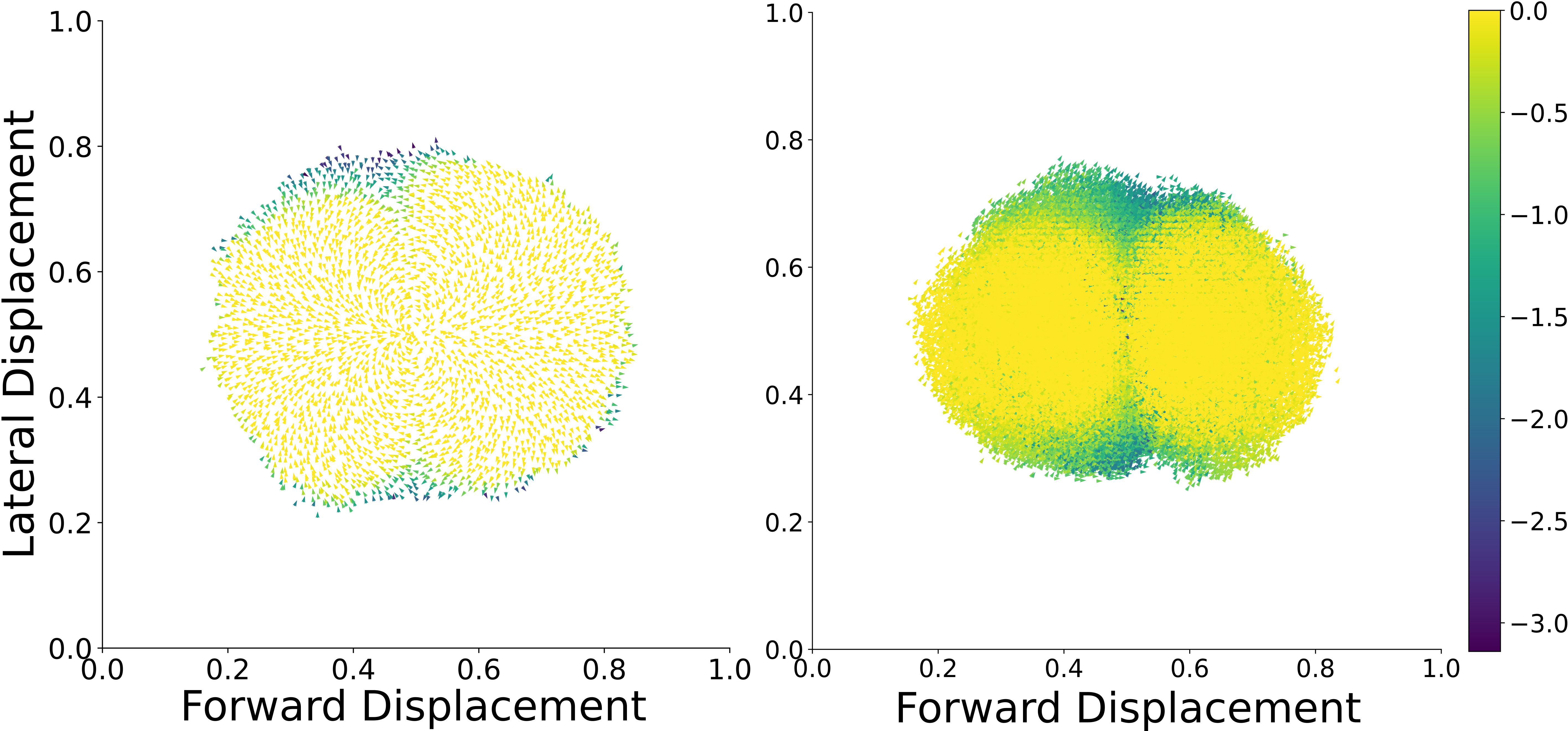}
\caption{The left image represents the repertoire that belongs to the \hbr{} method and the right image corresponds to the 8-dimensional flat variant. Both show the $x$ and $y$ location after 3 seconds of movements for the robot on a flat plane. The right repertoire has a higher density since we collapse 8 dimensions onto 2 to compare them. The legend shows the fitness of the solutions (i.e. robot's orientation).}
\label{fig:archives}
\end{figure}

%  To test whether the hierarchical stacking of repertoires helps breaking down the dimensionality of the optimisation problem at hand, we will compare our hierarchical approach against two "flat" archives. 

 In the following experiments we consider three baselines, namely RTE with a 2-dimensional repertoire (2D-RTE) \cite{Chatzilygeroudis2018Reset-freeRecovery}, RTE with a 8-dimensional repertoire (8D-RTE) and APROL \cite{Kaushik2020AdaptiveRobotics}. The "flat" repertoires we use for RTE should have the same solution space (and same behavioural space) as the HBR used in \hbr{}. The open-loop controllers for the flat repertoires are defined by a genotype with 36 parameters and they are executed for 3 seconds. For this purpose, the flat 8-dimensional repertoire version will have a behavioural descriptor space of $2+6$ dimensions where each controller is executed for 3 seconds. The first 2 dimensions correspond to the final $x$, $y$ displacement of the robot, while the other 6 dimensions represent a binary leg contact similarly to the middle layer of \hbr{}. For the 2-dimensional version, we use the final $x$, $y$ displacement as behavioural descriptor which is the same we use for the top layer of \hbr{} (see Fig. \ref{fig:archives}).
 
For the baselines, we use the code from  the original authors of RTE, and we re-implemented APROL in C++ with the SferesV2 framework \cite{Mouret2010}. 
As planning method within APROL, we use MCTS and A* similarly to RTE which differs from the original implementation where only A* was used. 
 
For APROL, we created 21 different flat 2-dimensional repertoires where each repertoire has been trained with a single or double leg damage on a flat terrain. For the flat repertoires used in RTE and APROL, we use the same hyper-parameters than in the original RTE paper.

For the simulations, we use the framework \emph{RobotDART} \footnote{\url{https://github.com/resibots/robot_dart}} which is an accessible C++ wrapper that allows to create robot tasks in the DART \cite{Lee2018} simulator. The implementations for HTE and the experiments are made available online\footnote{\url{https://github.com/adaptive-intelligent-robotics/HTE}}.% at [REDACTED]. 
 %\url{https://github.com/adaptive-intelligent-robotics/HTE}.
To compare the different repertoires, we created 4 repertoires for each variant and replicated the evaluations 30 times for each damage.

\subsection{Can the hierarchical repertoire be more diverse than a regular flat repertoire?}

 \begin{table}
\centering
 \begin{tabular}{c | c || c || c} 
 \toprule
    Algorithm & Repertoire Size & Effective Size & Mean Fitness  \\ [0.5ex] 
 \midrule
 \hbr{} & $2980.8 \pm 131.0$ & $2980.8 \pm 131.0 $&$-0.01 \pm 0.02 $\\
 
Flat-8D & $72528.0 \pm 2809.4$ &$2531.5 \pm 111.8$ &$-0.21 \pm 0.02 $\\
 \bottomrule
\end{tabular}
\caption{Results for each architecture (at $4x10^6$ evaluations). The numbers  of solutions in the repertoire differ by one order of magnitude since the flat repertoire has 8 dimensions and the HBR version only has 2. The \textit{Effective Size} is the number of cells we fill when projected on the first two dimensions and the \textit{Mean Fitness} is the mean fitness of these individuals.}
\label{table:hbr_results}
\end{table}

We want to compare the diversity of solutions produced by HBR, compared to the one produced by the 8D repertoire.
We present the results in Table~\ref{table:hbr_results}, where we observe that the flat version is able to find a high number of solutions. This result is expected since we have $640 000$ cells that can be filled. However, when projecting the content of the repertoire on the two first dimensions, namely the $x$, $y$ displacement, we can observe in Fig.~\ref{fig:archives} that \hbr{} covers a larger space (effective size) than the flat 8D repertoire. 
We report the coverage of this projection in Table \ref{table:hbr_results} as the \textit{Effective Size} which represents the number of cells that we have filled.
%We can observe that MAP-Elites is able to fill more cells on these two dimensions for the \hbr{} variant since there are only two dimensions along which we optimise the top layer of \hbr{}.
In these comparisons, \hbr{} has a better "effective" coverage and a better average fitness than the flat 8D repertoire. This demonstrates that \hbr{} can generate a better repertoire for the x, y displacements which are the skills we are interested in for the navigation tasks.
% One advantage of breaking down the behavioural dimensions into the hierarchy, is that we can guide the evolutionary algorithm more clearly towards wanted behaviours (i.e. x,y locomotion) whereas the flat version is trying to diversify towards all the dimensions of the defined behavioural space.
% Since we cannot push the additional 6 behaviours towards an external optimisation process as we do in \hbr{}, we are forced to optimise everything together with the consequence of losing diversity along certain dimensions.

\begin{figure}
\centering
    \begin{subfigure}[b]{0.45\textwidth}
    \centering
    \includegraphics[width=\textwidth]{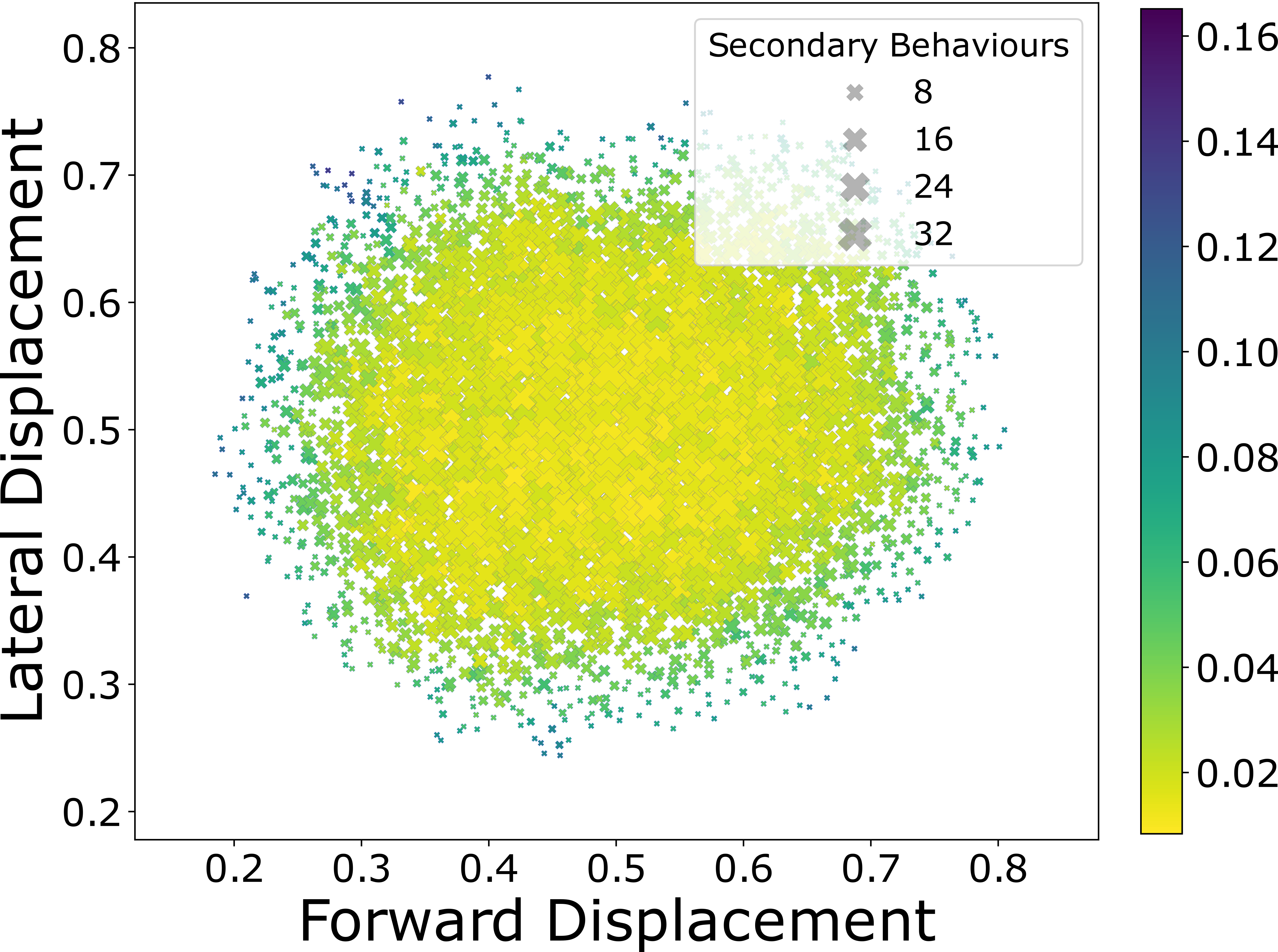}
    \caption{Final forward and lateral displacements ($x$, $y$ position) the hexapod robot can reach while using different secondary behavioural descriptors. The size of the crosses represents the number of secondary behaviours from the middle layer that we can execute without changing the behavioural descriptor of the solution in the top-layer repertoire. The colour represents the median error between the desired $x$, $y$ position and the robot's final position.}
    \label{fig:conditional_behaviours}
    \end{subfigure}
\hfill
\begin{subfigure}[b]{0.45\textwidth}
    \centering
    \includegraphics[width=\textwidth]{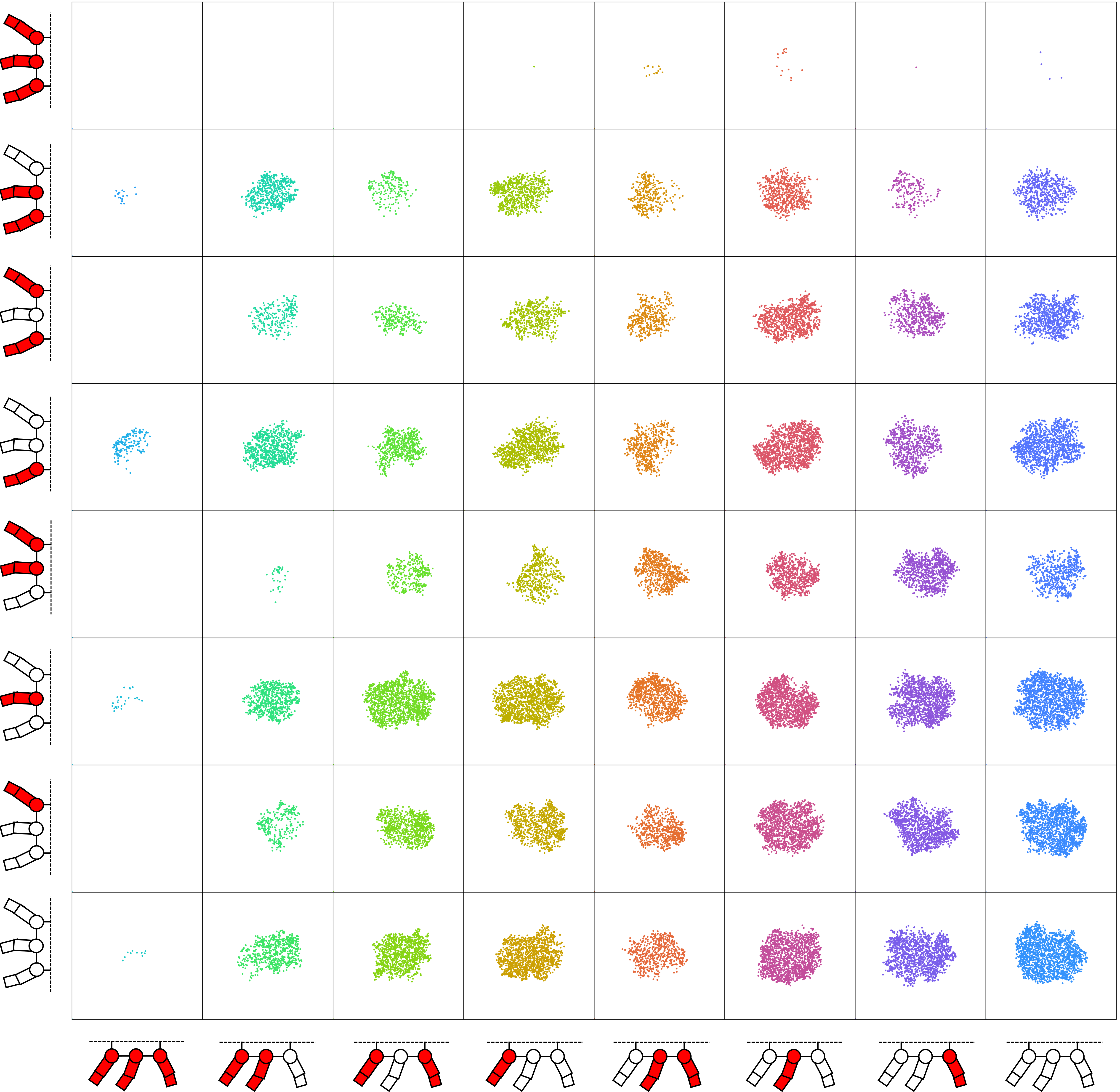}
    \caption{Each subplot shows the final $x$, $y$ positions we can reach with the hexapod while using different leg duties (i.e. more or less than 30\% contact with the ground). On this plot we have one subplot per desired secondary behavioural descriptor where each leg is red if it is not used for that subplot.}
    \label{fig:conditional_behaviours_grid}
    \end{subfigure}
\caption{Diversity Created with Secondary Behaviours}
\end{figure}

\subsection{Can we modulate high-level skills with secondary behavioural descriptors?}
To test whether we can modulate high-level skills, we re-evaluate all solutions from the top-layer of \hbr{} with the objective to reach the same behaviour descriptor (i.e. $x$ and $y$ position) while using different secondary behavioural descriptors in the middle-layer. 
For testing purposes, we use the same secondary behavioural descriptor for the full duration of the skill (i.e. the secondary behaviour will be the same for the 3 steps) which leaves us with 64 (2 choices per leg which gives us $2^6$ possible combinations) possible choices for each high-level skill. Some of these secondary behaviours are expected to be impossible to achieve for the robot (e.g. to not use any of its legs). After re-evaluating the solutions, we count how many times the robot (i) exhibited the wanted secondary behaviour and (ii) whether the robot was able to reproduce its behaviour descriptor by reaching its original final $x$, $y$ position.

In Fig.~\ref{fig:conditional_behaviours}, we plot how many times the robot reaches the desired $x$, $y$ location after 3 seconds while executing different secondary behavioural descriptors. The colour of the legend tells us how close to a desired $x$, $y$ location we get in the median case for executed movements that match the desired secondary behavioural descriptor. The size of the markers is proportional to the numbers of high-level skills that can be executed with secondary descriptors. On Fig.~\ref{fig:conditional_behaviours}, we have a big region of low error in the centre with a large numbers of valid secondary descriptors. Therefore, the robot is able to move to various $x$, $y$ positions while using its legs differently to get there. 
We can observe that the error grows and the number of secondary behaviours decreases near the edges of the repertoire. This inverse correlation can be expected as it is likely more challenging for the robot to find a large number of different ways to walk fast in different directions.

In Fig.~\ref{fig:conditional_behaviours_grid} each subplot corresponds to the final $x$, $y$ positions reachable with a specific secondary behavioural descriptor (64 in total). This result shows which secondary descriptors are actually not feasible (e.g. the first column or first row of the subplots). 
On both figures, we can observe that \hbr{} is able to use secondary behavioural descriptors to modulate its high-level skills when moving to the same final point. This show that \hbr{} can exhibit a different behaviour even though we have not expanded the behavioural space on the top layer during training. These results show that \hbr{} successfully reuses the diversity within the hierarchical structure.

\begin{figure}[h]%
\centering
\includegraphics[width=0.6\textwidth]{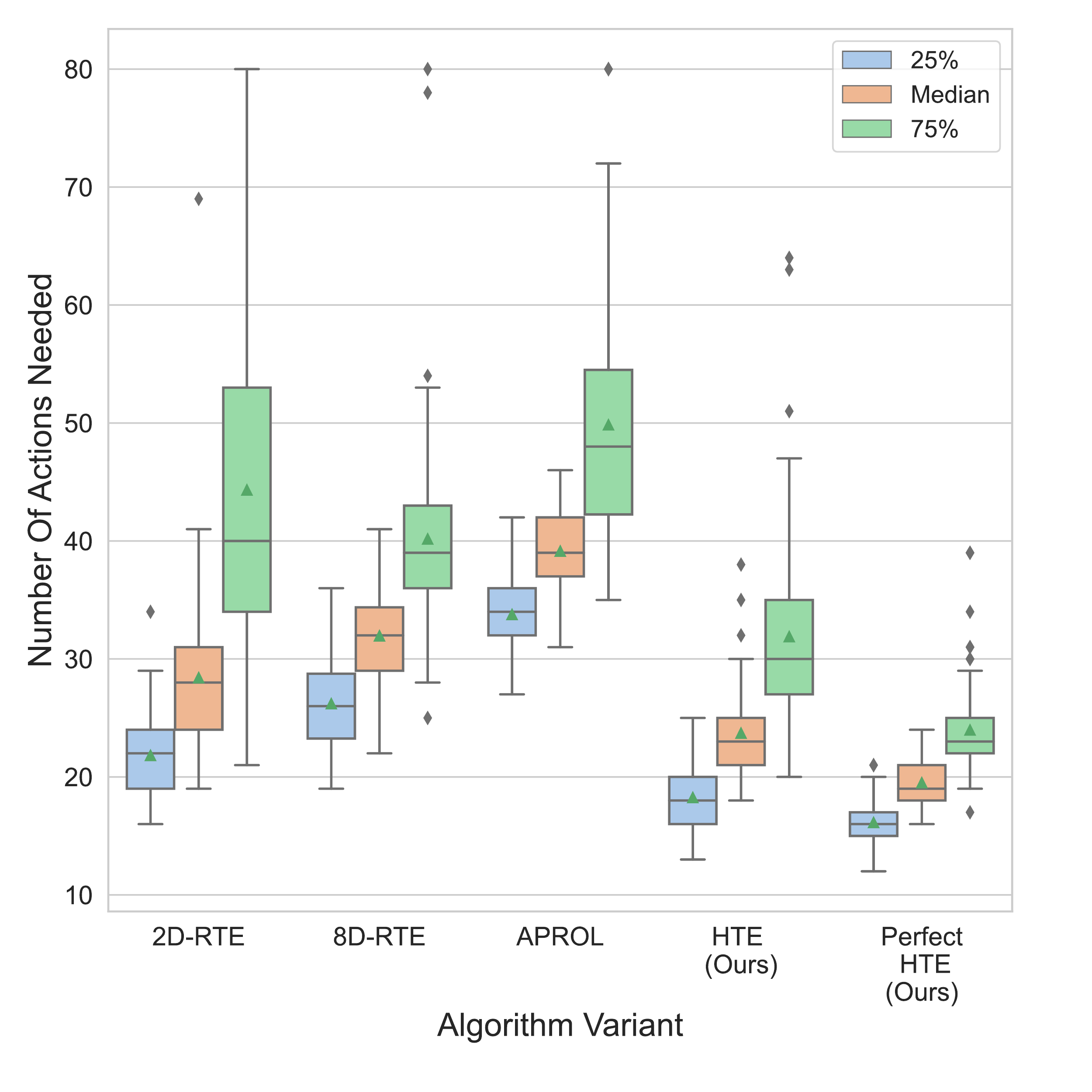}
\caption{Box-Plot for the different variants of algorithms for which we report the 25, 50 and 75 percentile values for each replication. \hbr{} is outperforming the baselines in each case showing that the hierarchy is helping with the adaptation to damages.}
\label{fig:actions_per_repertoire}
\end{figure}

\subsection{What are the benefits of \hbr{} on the robot's adaptation capabilities?}
Finally, we evaluate the benefits of \hbr{} and the additional behavioural diversity provided by the secondary behavioural descriptors on the robot's adaptation capabilities. More specifically, we test the algorithm in a maze navigation task with a damaged hexapod (see Fig.~\ref{fig:environment}).

We run \hbr{} in two different scenarios. First, an upper-baseline scenario where the damage of the robot is known and we can directly select a preferred secondary behaviour (called Perfect \hbr{}). Second, a scenario where we let the robot adapt automatically by choosing the optimal secondary behaviours for the given situation with Bayesian Optimisation (see Sec. \ref{sec:trial}).
We are interested in knowing which version is the most adaptive across different scenarios. Thus, we report the distribution of the median numbers of actions that a single variant replications needs across the 7 tasks. Since we did 30 replications per repertoire we obtained $7*30=210$ median values for our report per variant. Moreover, we are interested in how the variants perform in the best-case and worst cases which is why we also report the $25th$ and $75th$ percentiles in addition to the medians.

From the results in Fig.~\ref{fig:actions_per_repertoire}, we can see that the best performing algorithm is \hbr{} with prior knowledge. The manual selection of the secondary descriptor based on the prior knowledge of the damage helps the robot to adapt since it will only use moves that do not use the damaged leg. However, we expect that this prior knowledge and expertise might not always be available. When this information is not available, \hbr{} still performs better than all the other variants, even if it has to infer from experience the optimal secondary behaviours for recovery.

\begin{figure}[h]%
\centering
\includegraphics[width=0.5\textwidth]{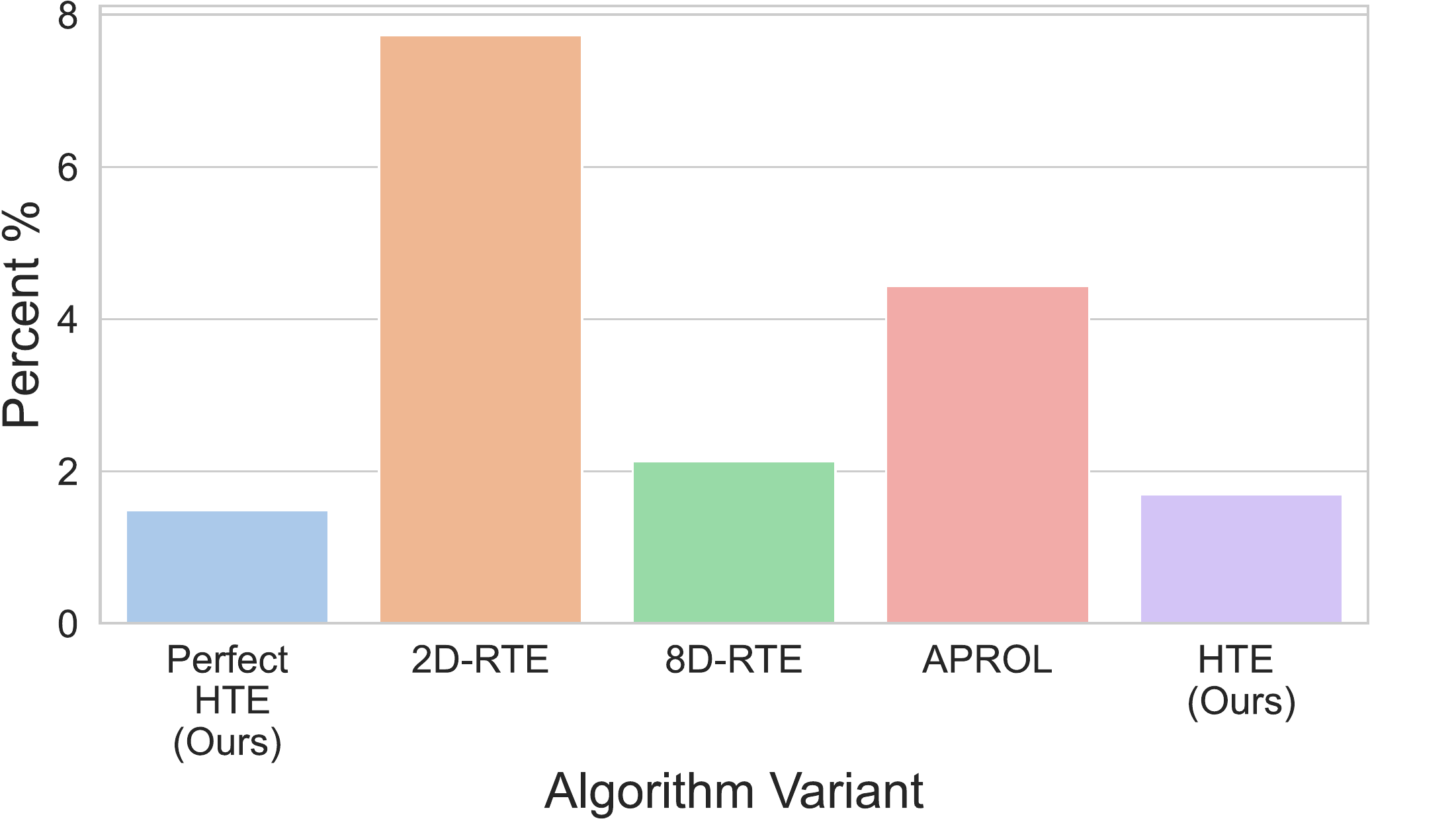}
\caption{Percentage of the experiments that failed to finish in under 80 actions for each variant.}
\label{fig:found_percentage}
\end{figure}

In median, we can observe across the replications that \hbr{} needs 24 actions, while 2D-RTE needs 28, 8D-RTE 32 and finally APROL $39$ actions. This result shows a $14.2\%$ improvement for \hbr{} over the best baseline, namely 2D-RTE for the median cases.
We can see from the results in Fig.~\ref{fig:actions_per_repertoire} that RTE with a 2D repertoire performs better than the other baselines in the 25th percentile range. For example, for the 25th percentile the robot only needs $22$ actions in median to solve the 7 tasks for the 2D-RTE variant and $19$ for the \hbr{} variant. APROL has the worst performance in the experiments which might be due to a reduced number of available repertoires (we use 21 repertoires instead of 57 repertoires in the original paper as we do not study the effect of friction). Another difference is that we use MCTS as a planner instead of A* from the original APROL study. For APROL, we see that even in the best scenarios we need $34$ actions and for 8D-RTE we need $26$.

\hbr{} performs well even in the worst case scenarios (i.e. the $75th$ percentiles) where it takes $32$ actions in median to solve the tasks in comparison to $40$ for the 2D-RTE, $39$ for 8D-RTE and $48$ for APROL. In these worst case scenarios, we show a $20\%$ improvement over the best baseline (2D-RTE). However, \hbr{} still fails to match the performance of the Perfect \hbr{} variant which only needs $16$ actions for the 25th percentile, $19$ for the median and $24$ for the $75th$ percentile of replications. Perfect \hbr{} shows that there is still room for the design of a better secondary behaviour selection in future work. 
The results on the downstream task for each algorithm are statistically different according to a Wilcoxon–Mann–Whitney test and a p-value < 1e-6 with a Bonferroni correction.

In a few cases, the robot may flip on its back and get stuck or stall in a corner of the maze. For these reasons, we limit the number of possible actions the robot can take to solve the navigation task to 80 (i.e. a failure). While this hard-coded value could bias the average of the results, the reported box-plots for the median values are less influenced by outliers. For the $25th$ and $75th$ percentile values, we do not use an interpolation for the same reasons, but we take the closest data point that corresponds to the $nth$-percentile value. 
In Fig.~\ref{fig:found_percentage}, we show the failures across replications of each variant for the task. \hbr{} has $78\%$ less failures in the maze experiments when compared to the best performing baseline in the maze (2D-RTE). 8D-RTE and APROL have less failures at the expenses of more steps. \hbr{} is able to both solve the maze quickly while being more stable with regard to total failures.

The results show that the diversity created by \hbr{} helps to improve the damage recovery capabilities of a hexapod robot in our maze experiment where the robot requires to find many diverse ways of executing a behaviour and navigate the maze.\\

\subsection{Is the Trial and Error method selecting useful secondary behaviours?}\label{sec:random_selection}

\begin{wrapfigure}{r}{0.49\textwidth}
  \begin{center}
    \includegraphics[width=0.4\textwidth]{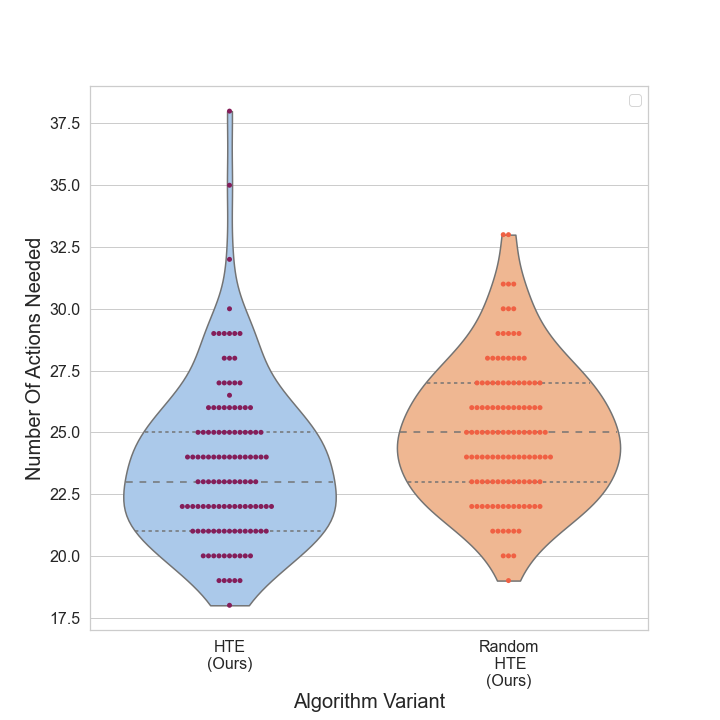}
  \end{center}
  \caption{Ablation of Selection Process for the Secondary Behaviours. The distributions of the median values for each repertoire show that selecting the secondary behaviours with the Trial and Error algorithm reduces the number of actions that are needed to solve the maze.}
\label{fig:gp_ablation}
\end{wrapfigure}
\hbr{} is effective at adapting to damages because of the added diversity through secondary behaviours in the middle layer. To evaluate the effect of the trial and error approach within \hbr{}, we ran \hbr{} with a random selection of secondary behaviours at every step. Random selection will not reduce the diversity of executed skills but it will show whether selecting secondary behaviours has an effect on the results. The results from Fig.~\ref{fig:actions_per_repertoire} suggest that choosing the right secondary behaviour in the perfect \hbr{} method enables our robot to solve the maze quicker than other approaches which is why we would like to get as close as possible to this perfect scenario. 
Fig.~\ref{fig:gp_ablation} shows the distribution of mean actions \hbr{} needs to solve the maze by using a random selection process in comparison to the trial and error approach described in Sec.~\ref{sec:trial}. The trial and error process is helping when solving the maze needs $23$ actions in the median case while the random selection requires $25$ actions to solve it. These results are statistically different according to a Wilcoxon–Mann–Whitney test and a p-value < 5e-4 with a Bonferroni correction.

\section{Adapting to Damages In The Physical World}

Additionally to the experiments presented above, we are interested in evaluating the capabilities of \hbr{} in the physical world. We want to evaluate the capability of our method to adapt to damages in the physical world. Simulators are usually not perfect representations of the physical world and can exhibit flaws which are then reflected in learnt policies. These flaws are the reason for a reality gap that can pose challenges and needs to be overcome by the algorithm. To this end, we use the repertoires that have been learned in simulation to solve a physical maze with a real damaged hexapod.  

\subsection{Experimental Setup}
The experimental setup consists of a 18-DoF hexapod robot and a maze (see Fig.~\ref{fig:real_maze}) which is almost identical to the maze used in simulation (see Fig.~\ref{fig:environment}).

The hexapod robot is controlled by 15 (5 legs x 3 motors) Dynamixel motors (type XM430-W350), three per leg. We removed a leg to simulate a scenario where the robot is damaged. The experiments are run with only a single damage to the right middle leg of the hexapod (see Fig.~\ref{fig:real_hexa}) in contrast to the 7 damages tested in simulation as a proof of concept in the real world. During the experiments, it is necessary to measure the observed behaviour of the robot, namely the position before and after a selected skill. For this purpose, we use a Vicon Motion Capture system to track the movements of the robots in our arena. The positions of the maze's walls are known to the robot during the deployment, which are  used by the MCTS planner to find the best path while avoiding collisions.

\hbr{} and the baselines (i.e. RTE and APROL) are identical to the ones used in simulation. We train 4 distinct repertoires per method and run 4 independent evaluations per repertoire (totalling 16 runs per algorithm). Each method is evaluated by counting the number of actions the robot needs reach the goal location (red cross in the lower left end of the mazes). In total, this requires $16x5 = 80$ experiments on the physical robot.

RTE, APROL and HTE use Monte-Carlo Tree Search (MCTS)~\cite{remi2006mcts} for the path planning within the maze. For these experiments, we use a parallel version of MCTS~\cite{cazenave2007} with 32 parallel roots and each tree having a budget of 100 iterations per planned action for every method. Having a fixed number of iterations per method will allow us to compare the time spent in the planning phase as well as the complexity of the search spaces (i.e. bigger skill search space needs more planning). We directly use an existing implementation of MCTS\footnote{\url{https://github.com/resibots/mcts}} in C++ for all our experiments.

\subsection{Can we adapt to robot damages in the physical world with policies learned in simulation?}

\begin{figure}[h]%
\centering
\includegraphics[width=1.0\textwidth]{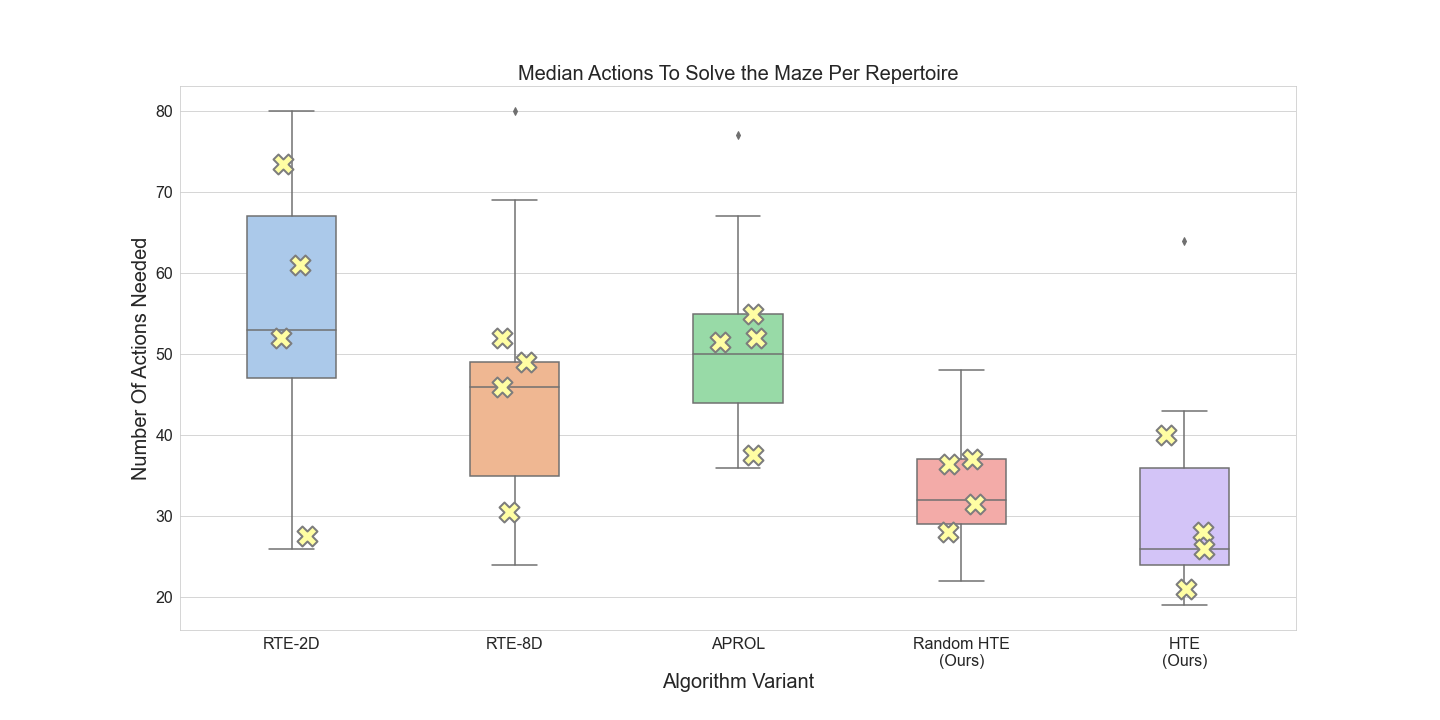}
\caption{Action distributions needed to solve the maze. The yellow crosses represent the median actions needed by the same repertoire. Since we have 4 repertoires per variant and we run the maze experiment 4 times per repertoire, we show both the distribution of all the actions (box-plots) and the distribution of actions per individual repertoires (crosses).}
\label{fig:real_maze_median_actions}
\end{figure}

Real-world applications do not only require robust actions but additionally they should be executed near real-time. For our experiments we use 32-Cores on a AMD Ryzen Threadripper 3970X 32-Core Processor machine to run the experiments with the robot. 
In addition to the baselines we have introduced earlier, we also include \hbr{} with a random secondary behaviour selection to analyse how our trial and error process works in real-world experiments similar to Sec.\ref{sec:random_selection}.

\paragraph{\textbf{Actions Needed To Solve the Maze.}} In Fig.~\ref{fig:real_maze_median_actions}, we show the distributions of the actions needed per variant to solve the maze in Fig.~\ref{fig:real_maze}. The first thing to note is that both Random \hbr{} and \hbr{} can solve the maze quicker than the other baselines. Among those two variants, \hbr{} needs $26$ median actions to solve the maze while Random \hbr{} needs $32$ actions. In contrast to the experiments done in simulation, it seems that RTE with a 2D repertoire is not able to adapt well to the damage while solving the maze ($53$ in median and $67$ for the $75th$ percentile). During the experiments, the RTE-2D method was not able to turn well to the left due to the missing right leg which resulted in more needed actions to solve the maze. In contrast to the 2D version, the RTE-8D version seemed to benefit from the additional diversity in the repertoire (using $46$ median actions). The additional diversity was helpful to overcome the damage, however having a larger skill space resulted in much longer execution times due to path planning with MCTS in the large skill space. 

\begin{wrapfigure}{r}{0.5\textwidth}
  \begin{center}
    \includegraphics[width=0.5\textwidth]{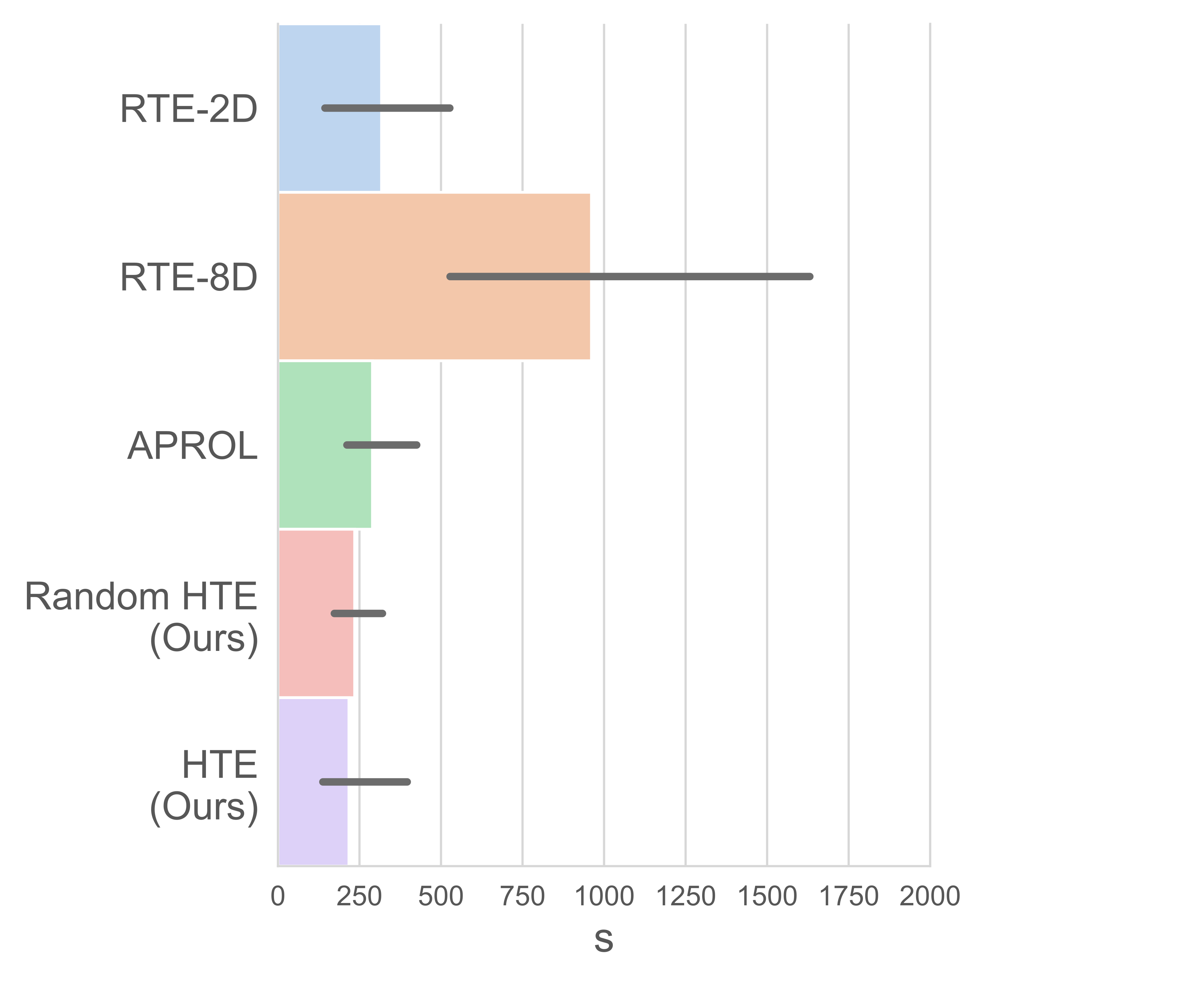}
    
  \end{center}
  \caption{Average seconds needed to solve the maze for each variant. Black bars represent the 95th percentile intervals. 8D-RTE is the slowest method due to its large skill space. \hbr{} is as quick as the 2D-RTE versions which shows that the hierarchical decomposition helps the planning algorithm to solve the maze faster.}
  \label{fig:real_maze_duration}
\end{wrapfigure}

\paragraph{\textbf{Deployment Time.}} Deployments in the real world benefit from quick adaptation and navigation. These benefits are the reasons why we are interested in the wall-time of each run in the physical world. Comparing execution times is subject to the hardware used during the run which is why we have the same conditions for each variant. To visualise the differences in speed for each algorithm, we plot the distributions of the duration (in seconds) in Fig.~\ref{fig:real_maze_duration} needed to solve the maze. 

The duration includes everything from planning to the execution of the controls on the robot. From the plot we can see that the increased diversity of skills in the 8 dimensional repertoires together with the MCTS planning causes the 8D-RTE variant to be slower than the other variants. 8D-RTE needs $300\%$ more time than RTE-2D and $440\%$ more time than \hbr{} in the mean case which is a considerable amount of time during a deployment. APROL benefits from distributing the diversity across different repertoires and is able to have results close to the RTE-8D variant in terms of actions while solving the task in less time. \hbr{} is the fastest method due to the reduced number of actions needed and the distributed diversity via the hierarchical organisation the repertoires. The hierarchical structure allows the path plannig to happen on a single layer (the top layer) which allows MCTS to finish as quick as 2D repertoires but while being able to adapt better (i.e. less actions needed to solve the maze) because of the increase diversity provided by secondary behaviours.

\paragraph{\textbf{Repertoire Dependencies.}} We can observe that the results depend on the trained repertoire to solve the maze fast. Repertoires that have learnt locomotion skills which do not use the damaged leg will be at an advantage in comparison to repertoires that have learnt locomotion skills that heavily depend on the damaged leg. This comes back to our assumption that repertoire-based adaptation needs to have learnt a solution which is not affected by the new situation. The more diversity we have in our repertoires, the better are the chances to adapt to an unforeseen scenario. In Fig.~\ref{fig:real_maze_median_actions}, we can observe the distributions of each variant across the repertoires. \hbr{} has three of the four repertoires that need less than 28 actions in median whereas the fourth repertoire seems to be less suited to the maze/damage combination we have evaluated on the real robot. This indicates that the diversity in the \hbr{} algorithm is good enough to overcome the maze task with damage. On the other hand, 2D-RTE in Fig.~\ref{fig:real_maze_median_actions} shows that an algorithm can be lucky during training and learn a repertoire with many well suited skills (i.e. a lot of skills do not use the damaged leg.) which only needs 28 median actions to walk through the maze but unlucky for others (all repertoires perform above 50 actions in median).

The results are statistically significant between all the variants and \hbr{}, with p<3e-2 calculated with a Wilcoxon–Mann–Whitney test Fig.~\ref{fig:real_maze_median_actions}. However, the difference between \hbr{} with Trial and Error and with random selection is not significant for the real-world experiments due to the restricted number of hardware experiments and the differences between the trained repositories.
Random \hbr{} is still quite effective in our experiments since there is a $50\%$ chance of guessing the right secondary behaviour (i.e each leg represents a binary choice). It is highly likely that with more choices, the Random \hbr{} variant will not find the best secondary behaviours and might perform worse in comparison to \hbr{}.

Finally, in our experiments  \hbr{} is not only able to adapt quicker in simulation but it can also transfer the learned skills to the physical world while still being $440\%$ quicker than the best baseline and $43\%$ less steps in the median case.

\section{Conclusions and Future Work}

In this paper, we introduced the Hierarchical Trial and Error algorithm, which allows robots to learn a diversity of skills with hierarchical repertoires and use them for damage recovery both in simulation and in the physical world. \hbr{} learns skills by splitting the complexity of the behavioural space across different layers. The introduction of primary and secondary behavioural descriptors allows \hbr{} to learn an additional diversity of skills during training while keeping the training tractable. The secondary behavioural descriptors make it possible for the repertoire to store a larger diversity of skills than other baselines without making the training process more complex. Furthermore, we showed that the trial and error process helps to take advantage of the diversity created by the secondary behaviours.

We can effectively use this diversity in an online fashion to recover from mechanical damages while solving a maze navigation task. The results from the simulation experiments show that the damage recovery by \hbr{} needs $14\%$ less actions than the best baseline and, more importantly, \hbr{} has less complete failures (i.e. needing more than 80 actions to solve the maze). Additionally in the real-world scenarios, \hbr{} requires even $43\%$ less actions than the best baseline while being over four times quicker in terms of execution time which is crucial for real-world experiments. In comparison to flat 2-dimensional repertoire variants, \hbr{} has similar execution times even though it offers more diversity thanks to the hierarchy. 
Finally, \hbr{} benefits from having a large diversity of behaviours while being fast in tasks requiring planning due to the reduced search space in the hierarchical structure.

One of the main limitations of this work is the definition of behavioural spaces on lower layers which heavily impacts the upper layers by restricting the search space. If the lower layers don't have enough diverse solutions, \hbr{} will not find optimal solutions for the upper layers. Consequently, \hbr{} requires task expertise to define different genotypes, behavioural descriptors and fitness functions. Finally, our results show that the perfect \hbr{} variant is better performing than \hbr{} which might indicate the there is room for improvement in the skill selection mechanism.

\begin{acks}
This work was supported by the Engineering and Physical Sciences Research Council (EPSRC) grant EP/V006673/1 project REcoVER. We want to also thank the members of Adaptive and Intelligent Robotics Lab (AIRL) for their valuable inputs.
\end{acks}

\bibliographystyle{ACM-Reference-Format}
\bibliography{references}

\end{document}